%% file: iclr2027_conference.tex
\documentclass{article} % For LaTeX2e
\usepackage{iclr2027_conference,times}

\input{math_commands.tex}

\usepackage{hyperref}
\usepackage{url}
\usepackage{svg}

\usepackage{subcaption}

\usepackage{booktabs} 
\usepackage{graphicx} % Required for inserting images
\usepackage{multirow}
\usepackage{multicol}
\usepackage{listings,textcomp}
\usepackage[T1]{fontenc}
\usepackage{lmodern}

\title{CorrGRPO: Correlation-Normalized GRPO for Multi-Reward Learning}

\author{\small Wenbin Hu\thanks{Equal Contribution}, Huihao Jing$^*$, Haochen Shi, Yuxuan Liu, Haoran Li,
Yangqiu Song \\
Hong Kong University of Science and Technology\\
\texttt{whuak@connect.ust.hk} \\
}

\iclrfinalcopy % Uncomment for camera-ready version, but NOT for submission.
\begin{document}

\maketitle

\begin{abstract}
Group Relative Policy Optimization (GRPO) is widely used to train reasoning language models, where it computes advantages by centering and normalizing rewards across rollouts of the same prompt. For multiple rewards, GRPO sums the reward components and normalizes the total reward by its within-group standard deviation. The corresponding variance equals the sum of all pairwise reward covariances. For a fixed centered reward, larger aggregate covariance produces smaller advantages, and vice versa, allowing update magnitudes to adapt to reward dependence. However, correlated rewards with large scales can dominate this normalization and suppress signals from smaller-scale rewards. We propose \textbf{Corr}elation-Normalized \textbf{GRPO} (\textbf{CorrGRPO}), which normalizes pairwise covariances into Pearson correlation coefficients. CorrGRPO keeps the centered total reward unchanged while balancing the influence of differently scaled rewards on the correlation-based normalization. This allows advantage magnitudes to adapt to reward correlations without the normalization being dominated by large-scale reward components. We compare CorrGRPO with GRPO and other variants on code generation, tool calling, and agent security, using models ranging from 0.5B to 8B parameters. These tasks all involve multiple rewards that can improve together or present tradeoffs. Results show improvements across three domains, including code generation, tool calling, and agent security. Our code is available at \url{https://github.com/HKUST-KnowComp/CorrGRPO}.
% and an outward expansion of the empirical Pareto frontier between competing reward objectives.
\end{abstract}

% \clearpage
% {
%   % \small
%   \setcounter{tocdepth}{2}
%   \tableofcontents
% }
% \clearpage

\input{latex/1-intro}

\input{latex/2-grpo-analysis}
\input{latex/3-corrgrpo}
\input{latex/4-exp}

\input{latex/5-related-works}

\input{latex/6-conclusion}

\bibliography{iclr2027_conference}
\bibliographystyle{iclr2027_conference}

\appendix

\clearpage
\input{latex/5-discussion}

\input{latex/7-appendix}

\end{document}

%% file: math_commands.tex
\usepackage{amsmath,amsfonts,bm}

\def\eqref#1{equation~\ref{#1}}
\def\1{\bm{1}}

\DeclareMathAlphabet{\mathsfit}{\encodingdefault}{\sfdefault}{m}{sl}
\SetMathAlphabet{\mathsfit}{bold}{\encodingdefault}{\sfdefault}{bx}{n}

%% file: latex/1-intro.tex
\begin{figure*}[h]
% \vspace{-0.2in}
    \centering
    % Left: formulas
    \begin{minipage}[c]{0.67\textwidth}
        \centering
        \includegraphics[width=\linewidth]{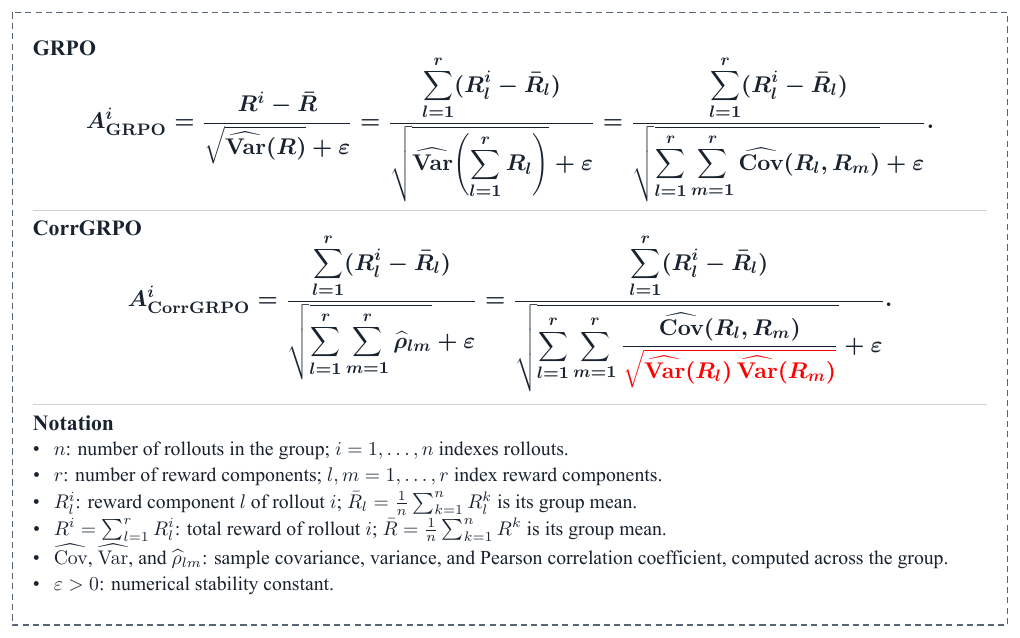}
    \end{minipage}\hfill
    % Right: accuracy above efficiency
    \begin{minipage}[c]{0.32\textwidth}
        \centering
        \includegraphics[width=\linewidth]{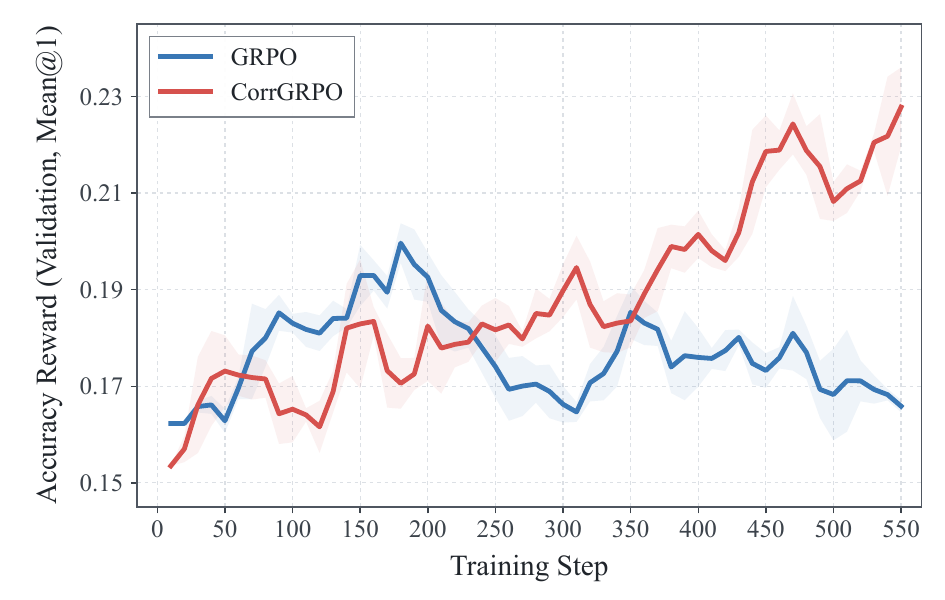}\par
        \vspace{0.4em}
        \includegraphics[width=\linewidth]{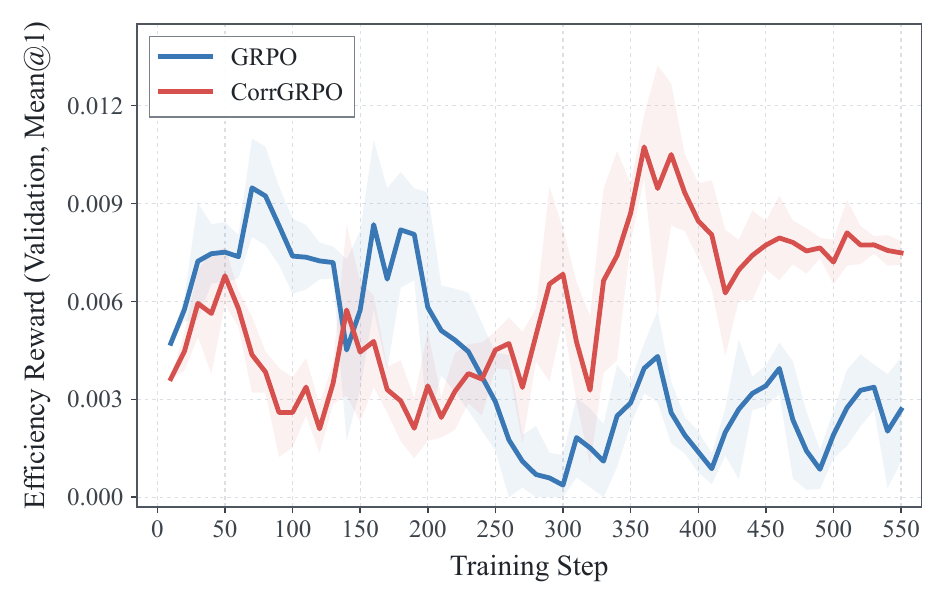}
    \end{minipage}
    \caption{
        An overview of CorrGRPO.
        (1) Left: CorrGRPO replaces total-reward standard deviation normalization with a correlation-based denominator while retaining the centered total reward.
        % the additional variance normalization is highlighted in red.
        (2) Right: Validation accuracy (top) and efficiency (bottom), measured by Mean@1, for Qwen2.5-Coder-7B-Instruct on LeetCodeDataset.
    }
    \vspace{-0.15in}
    \label{fig:grpo_corrgrpo_comparison}
    \vspace{-0.0in}
    
\end{figure*}

\section{Introduction}
% Introduction fragment for insertion into the manuscript.
% Requires amsmath, amssymb, and natbib; bibliography entries are in introduction_refs.bib.
% Figure 1 is referenced literally because the manuscript's figure label was not supplied.
\label{sec:introduction}

Reinforcement learning (RL) has become a central paradigm for aligning large language models (LLMs) with human preferences and improving their ability to solve complex tasks. By constructing training environments and designing rewards that capture desired outcomes, RL enables models to improve through feedback on their own generations, supporting both instruction following and the development of reasoning capabilities \citep{ouyang2022training,deepseekai2025deepseekr1}. Among existing approaches, Group Relative Policy Optimization (GRPO) has gained widespread adoption for its simplicity and efficiency. GRPO estimates advantages by subtracting the mean reward and dividing by the reward standard deviation within a group of responses sampled for the same prompt, eliminating the need for a separate value model \citep{shao2024deepseekmath}.

Practical training objectives often require multiple rewards to capture different aspects of desirable behavior. First, reward objectives can reinforce one another. Code generation, for example, can combine rewards for executability, test-case pass rate, and abstract syntax tree (AST) similarity to reference solutions; resolving syntax or runtime errors can improve both executability and test-case performance. Second, reward objectives can introduce tradeoffs. Agent training requires balancing task utility with security, where an overly conservative agent may avoid malicious instructions by also refusing legitimate requests, while an overly permissive agent may complete more tasks at the cost of greater exposure to prompt injection \citep{debenedetti2024agentdojo}. Such settings involve rewards that may improve together or exhibit tradeoffs. Their statistical relationships therefore provide a useful perspective on how multiple feedback signals interact during learning. In this work, we investigate how correlations among reward components affect advantage estimation in GRPO.

Our starting point is a simple identity with direct implications for multi-reward optimization. When GRPO aggregates $r$ reward components into a total reward $R=\sum_{l=1}^{r}R_l$, the variance underlying its normalization is
\(
\widehat{\operatorname{Var}}\!\left(\sum_{l=1}^{r}R_l\right)
=
\sum_{l=1}^{r}\sum_{m=1}^{r}
\widehat{\operatorname{Cov}}(R_l,R_m).
\)
The underlying population identity is derived in Appendix~\ref{app:corrgrpo_population_identity}.
Thus, the denominator depends on the sum of all entries in the reward covariance matrix, incorporating both individual reward variances and pairwise dependencies. Holding the marginal reward variances and centered total reward fixed, stronger positive correlations reduce the advantage magnitude by increasing the normalization denominator; weaker or more negative correlations increase the advantage magnitude by decreasing the denominator. GRPO therefore implicitly adjusts advantage scaling according to reward correlation within each rollout group. This provides a dynamic mechanism for modulating the strength of the aggregate learning signal as relationships among rewards change.

However, this mechanism entangles reward dependence with reward scale. Each covariance satisfies $\operatorname{Cov}(R_l,R_m)=\sigma_l\sigma_m\rho_{lm}$, where $\sigma_l$ and $\sigma_m$ are the component standard deviations and $\rho_{lm}$ is their Pearson correlation coefficient. Consequently, reward components with large scales of variation can dominate both the diagonal and off-diagonal terms in the covariance sum. The normalization then becomes disproportionately sensitive to these components, while smaller-scale rewards contribute little to the adjustment. This imbalance suppresses the influence of smaller-scale rewards on normalization, even when they encode useful distinctions among responses. This limitation is particularly relevant when heterogeneous rewards differ substantially in their numerical ranges or within-group variability.

To address this issue, we propose {Correlation-Normalized GRPO (CorrGRPO)}. As shown in Figure~\ref{fig:grpo_corrgrpo_comparison}, CorrGRPO replaces pairwise covariances in the denominator with Pearson correlation coefficients while retaining the centered total reward in the numerator. Normalizing each covariance by the corresponding component standard deviations removes its dependence on reward scale. Each reward component with nonzero within-group variance consequently contributes equally to the diagonal of the correlation matrix, while off-diagonal terms reflect the strength and direction of pairwise dependence. The denominator remains responsive to reward correlations without being dominated by components solely because they vary on larger numerical scales. Meanwhile, preserving the numerator retains the relative reward weights specified by the training objective. 
As a result, CorrGRPO allows correlations involving smaller-scale rewards to influence advantage normalization without being downweighted by their scales.
% while preserving the original reward preferences in the numerator.

We conduct experiments with CorrGRPO on code generation, tool calling, and agent security, three domains that naturally require multiple reward signals, using models ranging from 0.5B to 8B parameters. Comparisons with GRPO and relevant variants show improvements in code accuracy and efficiency, tool-call accuracy, and the balance between agent utility and security. Together, these results support correlation-based normalization as an effective approach to improving practical multi-reward learning.

Our contributions are threefold:
\begin{enumerate}
    \item \textbf{A covariance perspective on GRPO.} We characterize how the reward covariance matrix implicitly controls advantage scaling in multi-reward GRPO and show that large-scale rewards can dominate this adjustment.
    \item \textbf{Correlation-normalized advantage estimation.} We introduce CorrGRPO, which replaces covariance-based normalization with correlation-based normalization while preserving the centered total reward, preventing large-scale rewards from disproportionately influencing the denominator.
    % and retaining adaptation to reward dependence.
    \item \textbf{Experiments across three multi-reward domains.} We conduct extensive experiments on code generation, tool calling, and agent security across models from 0.5B to 8B parameters, demonstrating improvements over GRPO and relevant variants in the three domains, along with an expansion of the empirical Pareto frontier between reward objectives.
\end{enumerate}

%% file: latex/2-grpo-analysis.tex
% Section fragment for insertion into the manuscript.
% Requires amsmath, amssymb, and graphicx in the main document preamble.
% Display math uses explicit 9pt type (ICLR small) inside a local math box.
% Equation numbers and surrounding prose retain their normal size.

\section{A Covariance Normalization View of Multi-Reward GRPO}
\label{sec:grpo_covariance}

Group Relative Policy Optimization (GRPO)~\citep{shao2024deepseekmath} estimates advantages using rewards from a group of trajectories, avoiding a separate value model. For each prompt $q$, it samples $n$ trajectories $\{\tau_i\}_{i=1}^{n}$ from an old policy $\pi_{\theta_{\mathrm{old}}}$. With $r$ reward components, let $R_l^i=R_l(q,\tau_i)$ and $R^i=\sum_l R_l^i$, with group means $\bar R_l=\frac1n\sum_i R_l^i$ and $\bar R=\frac1n\sum_i R^i$. Any fixed reward weights are absorbed into the corresponding components. Under outcome supervision, all generated tokens in trajectory $i$ share the same advantage $A_{\mathrm{GRPO}}^i$. The advantage is computed by centering the total reward and normalizing it by its within-group standard deviation: 
\(A_{\mathrm{GRPO}}^i
=
\frac{R^i-\bar R}
{\sqrt{\widehat{\operatorname{Var}}(R)}+\varepsilon}\)
% \begin{equation}
% \mbox{\fontsize{9}{10}\selectfont$\displaystyle
% A_{\mathrm{GRPO}}^i
% =
% \frac{R^i-\bar R}
% {\sqrt{\widehat{\operatorname{Var}}(R)}+\varepsilon},
% $}
% \label{eq:grpo_advantage}
% \end{equation}
, where $\varepsilon>0$ ensures numerical stability. 
% The reinforcement learning setup and clipped surrogate objective are provided in. 
% We provide detailed RL setup in Appendix~\ref{app:rl_background}.

Our observation is that, with multiple reward components, GRPO implicitly uses the aggregate reward covariance to control advantage scaling. Applying the variance-of-a-sum identity, as derived in Appendix~\ref{app:corrgrpo_population_identity}, we obtain
\begingroup
% Compact display spacing, scoped to this equation only.
\setlength{\abovedisplayskip}{5pt plus 1pt minus 1pt}
\setlength{\belowdisplayskip}{5pt plus 1pt minus 1pt}
\setlength{\abovedisplayshortskip}{0pt plus 1pt}
\setlength{\belowdisplayshortskip}{3pt plus 1pt minus 1pt}
\begin{equation}
\mbox{\fontsize{9}{10}\selectfont$\displaystyle
A_{\mathrm{GRPO}}^i
=
\frac{R^i-\bar R}
{\sqrt{\widehat{\operatorname{Var}}(R)}+\varepsilon}
=
\frac{\displaystyle\sum_{l=1}^{r}(R_l^i-\bar R_l)}
{\displaystyle\sqrt{\widehat{\operatorname{Var}}
\left(\sum_{l=1}^{r}R_l\right)}+\varepsilon}
=
\frac{\displaystyle\sum_{l=1}^{r}(R_l^i-\bar R_l)}
{\displaystyle\sqrt{
\sum_{l=1}^{r}\sum_{m=1}^{r}
\widehat{\operatorname{Cov}}(R_l,R_m)
}+\varepsilon}.
$}
\label{eq:grpo_covariance_decomposition}
\end{equation}
\endgroup
Here, $\widehat{\operatorname{Var}}$ and $\widehat{\operatorname{Cov}}$ are computed across the same rollout group with a consistent normalization convention.
% Population and exact finite-sample derivations are provided in Appendices~\ref{app:corrgrpo_population_identity} and~\ref{app:corrgrpo_finite_sample}, respectively.
This decomposition reveals an implicit mechanism in GRPO: the denominator adjusts advantage scaling through both individual reward variances and pairwise reward dependencies.

\begin{figure*}
    \centering
    \includegraphics[width=0.98\linewidth]{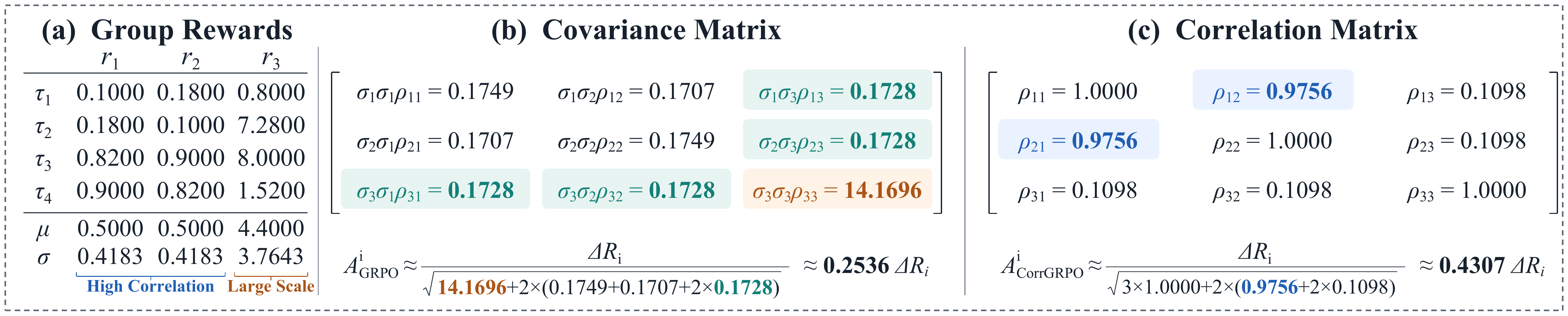}
    \caption{
An example for comparing GRPO and CorrGRPO.
\textbf{(a)} A group of four trajectories with three rewards. Rewards $r_1$ and $r_2$ are highly correlated, while $r_3$ has a larger scale and weak correlations with both.
\textbf{(b)} Covariance-matrix elements. In GRPO, the large scale of $r_3$ dominates the denominator.
\textbf{(c)} Pearson correlation coefficients matrix elements. 
% CorrGRPO removes reward-scale dependence from the denominator while preserving the numerator. 
In CorrGRPO, the strong correlation between $r_1$ and $r_2$ has a greater influence on the normalization denominator than $r_3$ does.
% Here, $\Delta R_i = R_i-\bar{R}$ denotes the centered total reward.
}
\vspace{-0.1in}
    \label{fig:example}
\end{figure*}

Specifically, let $\Delta R^i=R^i-\bar R$ and $S=\sum_{l,m}\widehat{\operatorname{Cov}}(R_l,R_m)$. The decomposition gives $A_{\mathrm{GRPO}}^i=g\Delta R^i$, where $g=(\sqrt S+\varepsilon)^{-1}$ is a scaling coefficient shared by all trajectories in the group. Positive covariances increase the variability of the summed reward, reducing this coefficient, while negative covariances offset part of that variability and increase it. Holding the marginal reward variances and a trajectory's centered total reward fixed, stronger positive correlations therefore reduce its advantage magnitude; weaker or more negative correlations increase it.
Because these statistics are recomputed from newly sampled trajectories, the scaling coefficient adapts to reward relationships across prompts and throughout training. This mechanism changes the scale of the group's reward-driven contribution to the surrogate objective while preserving the relative reward weights in the numerator. 
% Our analysis isolates this normalization effect by holding the centered total reward fixed.

However, covariance couples statistical dependence with reward scale. Each pairwise correlation is weighted by the product of the component standard deviations: $S=\sum_{l=1}^{r}\hat\sigma_l^2+2\sum_{l<m}\hat\sigma_l\hat\sigma_m\hat\rho_{lm}$, where $\hat\sigma_l^2=\widehat{\operatorname{Var}}(R_l)$. Large-scale rewards can dominate both the variance terms and the sensitivity of the denominator to changes in correlation. Consequently, the shared scaling coefficient can be driven primarily by these components, while dependencies among smaller-scale rewards have limited influence on the adjustment. 

Figure~\ref{fig:example} illustrates this scale imbalance with four trajectories and three rewards. The first two rewards are strongly correlated ($\hat\rho_{12}\approx0.9756$), while the third has weak correlations with both ($\hat\rho_{13}=\hat\rho_{23}\approx0.1098$). Nevertheless, the third reward's variance, $14.1696$, alone accounts for approximately $91.1\%$ of the covariance sum $S=15.5520$. Its covariances with the other rewards, both $0.1728$, also exceed the covariance $0.1707$ between the strongly correlated first two rewards. Thus, the third reward's large scale dominates the normalization despite its weak correlations, motivating CorrGRPO's removal of component-scale effects from the pairwise normalization terms.

%% file: latex/3-corrgrpo.tex
% Requires natbib. Add compatibility_references.bib to the bibliography.
% Revised Section 3: group-gradient scaling and a shared correlation metric.
% Include appendix_corrgrpo_analysis_v6.tex after the existing \appendix command.
% References to Section 2, Equation (2), and Figure 2 follow the supplied PDF.
% Pearson correlations below assume nonzero component variances. Document
% the actual implementation's treatment of constant components separately.

\section{CorrGRPO: Correlation-Normalized GRPO}
\label{sec:corrgrpo}

Motivated by the scale imbalance identified in Section~\ref{sec:grpo_covariance}, we propose
Correlation-Normalized GRPO (CorrGRPO), 
% The key idea is to normalize each
% pairwise covariance by the corresponding component standard deviations before
% aggregating it into the advantage denominator. This allows reward dependence
% to guide normalization without being disproportionately influenced by
% large-scale components.
which replaces the covariance terms in GRPO's denominator with sample
Pearson correlation coefficients:
\begin{equation}
A_{\mathrm{CorrGRPO}}^i
=\frac{\sum_{l=1}^{r}(R_l^i-\bar{R}_l)}
{\sqrt{\sum_{l=1}^{r}\sum_{m=1}^{r}\hat{\rho}_{lm}}+\varepsilon},
\qquad
\hat{\rho}_{lm}
=\frac{\widehat{\mathrm{Cov}}(R_l,R_m)}
{\sqrt{\widehat{\mathrm{Var}}(R_l)\widehat{\mathrm{Var}}(R_m)}}.
\label{eq:corrgrpo_advantage}
\end{equation}
All statistics are computed within the same rollout group, following
Section~\ref{sec:grpo_covariance}. 
For zero-variance reward components, we set the corresponding rows and columns of the correlation matrix to zero, including diagonal entries, as shown in the core implementation in Appendix~\ref{sec:corrgrpo-implementation}. 
The resulting matrix remains positive semidefinite, ensuring a nonnegative correlation sum and a strictly positive denominator when \(\varepsilon>0\), with derivation in Appendix~\ref{app:denominator_positivity}.
% We substitute this advantage into the clipped surrogate objective in Equation~\ref{eq:grpo_surrogate}. 
% The centered total reward is retained in the numerator, preserving the relative reward weights specified by the training objective.

Replacing covariances with correlations removes reward-scale weighting from each pairwise term. Each reward
contributes one to the diagonal, while off-diagonal entries reflect the
strength and direction of reward dependence. Consequently, strong correlations
between small-scale rewards can substantially influence normalization even
when other components have much larger variances. 
% This addresses the imbalance in Section~\ref{sec:grpo_covariance}, where these relationships can be overshadowed by weak correlations involving large-scale rewards. \\
% CorrGRPO also retains adaptation to reward relationships. For a fixed centered
% total reward, stronger positive correlations increase the denominator and
% reduce advantage magnitude; weaker or more negative correlations decrease the
% denominator and increase it. Recomputing these correlations for each group
% allows the scaling to adapt across prompts and throughout training. Together
% with the unchanged numerator, this design preserves the specified reward
% tradeoffs while allowing dependencies among differently scaled rewards to
% participate in determining the strength of the group's learning signal.
For example, in Figure~\ref{fig:example}, CorrGRPO assigns a correlation term of $0.9756$ to the strongly correlated rewards $r_1$ and $r_2$, compared with $0.1098$ for each weak relationship involving $r_3$. The stronger correlation contributes approximately $8.89$ times as much to the sum inside the denominator. 
% All diagonal entries become one, so the normalization reflects reward correlations without being dominated by component scales.

%% file: latex/4-exp.tex
% Section 4 opening and Section 4.1 only; insert into the manuscript.
% Requires amsmath, amssymb, and natbib (provided by the ICLR template).
% Merge section4_coding_tools_references.bib into the manuscript bibliography.
% Table 1 label: tab:qwen25_combined_results.
% Add \label{fig:coding_pareto} after the existing Figure 3 caption.
% The replacement Figure 3 caption is provided at the end as a commented block.
% RL hyperparameters are script defaults, not verified overrides for every table row.
% Epoch count omitted: saved versions specify both 3 and 15 epochs.
% Provenance: numeric weights below follow the saved scalar-reward configuration.
% Their use in individual CorrGRPO runs remains to be checked against run configs.

% \newcommand{\resulttablefontsize}{\fontsize{8}{9.6}\selectfont}
\newcommand{\resulttablefontsize}{\scriptsize}
% To override one table, replace its \resulttablefontsize below with a size command.
\newcommand{\securitytablefontsize}{\fontsize{8}{9}\selectfont}
\newcommand{\codertablefontsize}{\fontsize{8}{9.6}\selectfont}
\newcommand{\toolcalltablefontsize}{\fontsize{8}{9.6}\selectfont}
% Required packages in the preamble: \usepackage{booktabs,multirow}

\input{latex/table-code-rl}

\section{Experiments}
\label{sec:experiments}

We evaluate whether CorrGRPO improves task performance when language models are trained with multiple reward signals. Our experiments cover code generation, tool calling, and agent utility and security. We compare CorrGRPO with GRPO \citep{shao2024deepseekmath} and the corresponding base models, and additionally include GDPO \citep{liu2026gdpo} in the coding experiments. We report both overall task performance and individual reward-related metrics to examine how CorrGRPO balances different objectives. Furthermore, we show our training dynamics in Figure~\ref{fig:reward_curves} and provide detailed training settings in Appendix~\ref{app:training_settings}.

\subsection{Coding Reasoning}
\label{sec:coding_reasoning}

\paragraph{Task.}
We study coding RL with Python code generation.
% where a model generates a program from a natural-language problem description. 
The goal is to produce functionally correct programs while improving execution efficiency. We evaluate generated programs using executable tests and compare the runtime of passing solutions with that of the reference implementations.

\paragraph{Reward design.}
Let $y_g$ denote the generated program and $y_r$ the reference implementation. We combine seven reward components:
\begin{equation}
\begin{aligned}
R_{\mathrm{code}}(y_g,y_r)
={}&0.05R_{\mathrm{fmt}}(y_g)
+0.05R_{\mathrm{syn}}(y_g)
+0.05R_{\mathrm{compile}}(y_g)
+0.25R_{\mathrm{run}}(y_g)\\
&+0.60R_{\mathrm{pass}}(y_g)
+0.30R_{\mathrm{ast}}(y_g,y_r)
+0.20R_{\mathrm{eff}}(y_g,y_r).
\end{aligned}
\label{eq:coding_reward}
\end{equation}
The format reward $R_{\mathrm{fmt}}$ checks whether the response follows the required code format. The execution-related rewards follow a dependency chain:
\[
\underbrace{\text{Syntax validity}}_{R_{\mathrm{syn}}}
\;\rightarrow\;
\underbrace{\text{Successful compilation}}_{R_{\mathrm{compile}}}
\;\rightarrow\;
\underbrace{\text{Runtime success}}_{R_{\mathrm{run}}}
\;\rightarrow\;
\underbrace{\text{All tests passed}}_{R_{\mathrm{pass}}}.
\]
\vspace{-0.1in}

\begin{figure}[t]
    \centering
    \includegraphics[width=0.8\linewidth]{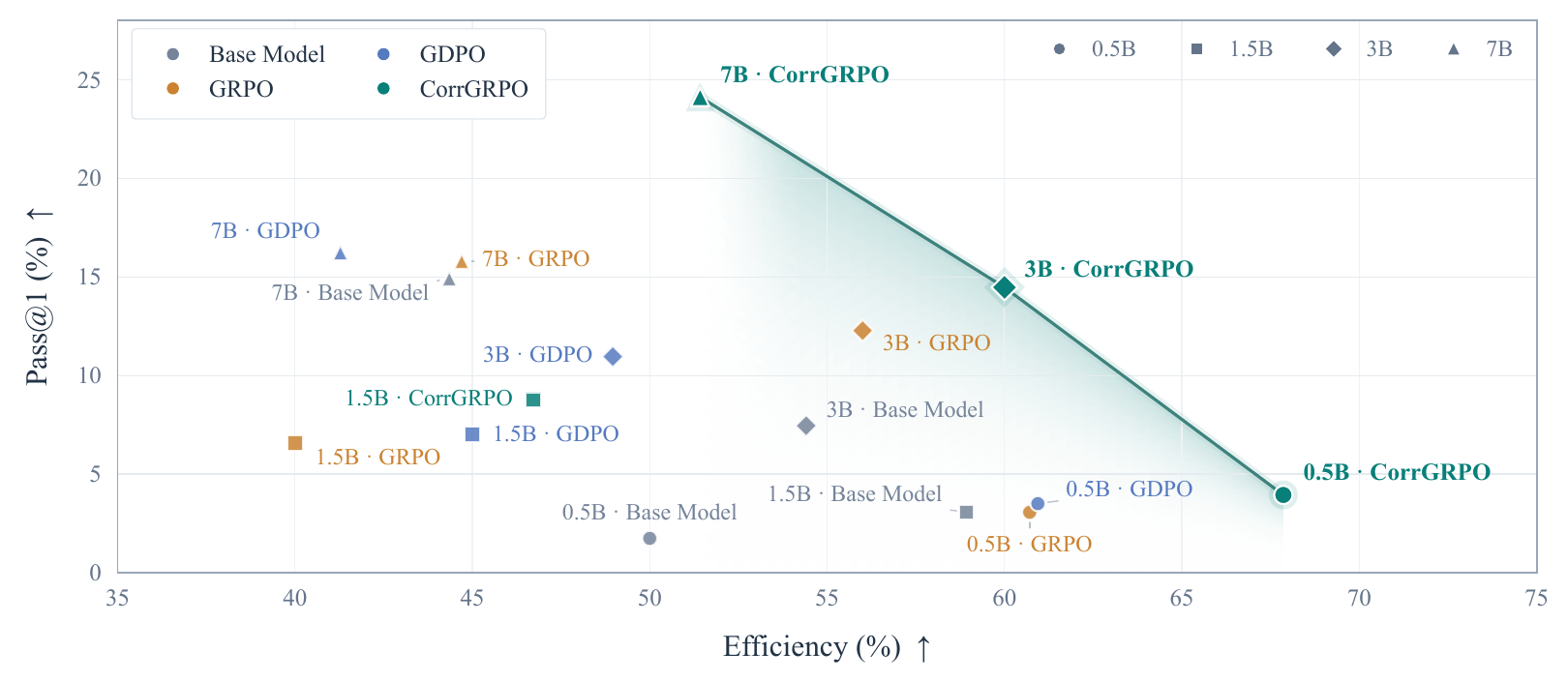}
    \vspace{-0.1in}
    \caption{Pareto frontier of correctness–efficiency tradeoff on LeetCodeDataset.
    % CorrGRPO models expand the frontier.
    }
    \vspace{-0.1in}
    
    \label{fig:pareto_frontier}
    \vspace{-0.1in}
\end{figure}

These rewards are binary and dependent: the compilation reward requires valid syntax, the runtime reward requires successful compilation, and the all-pass reward requires execution without runtime errors. Improving earlier checks therefore enables rewards from subsequent checks. Given the set of test cases $\mathcal C$, the final correctness reward is
\(R_{\mathrm{pass}}(y_g)
=\prod_{c\in\mathcal C}
\mathbf{1}\!\left[y_g\text{ passes }c\right].\)
% \begin{equation}
% R_{\mathrm{pass}}(y_g)
% =\prod_{c\in\mathcal C}
% \mathbf{1}\!\left[y_g\text{ passes }c\right].
% \label{eq:coding_all_pass}
% \end{equation}
Thus, $R_{\mathrm{pass}}(y_g)=1$ only when every test passes, while syntax, compilation, and runtime rewards provide intermediate feedback even when $R_{\mathrm{pass}}(y_g)=0$.
The structural reward $R_{\mathrm{ast}}(y_g,y_r)\in[0,1]$ measures the similarity between the generated and reference abstract syntax trees, independently of identifier names and literal values. Appendix~\ref{app:ast_similarity} describes the tree representation and similarity computation.
Finally, the efficiency reward encourages faster execution among functionally correct programs:
\(R_{\mathrm{eff}}(y_g,y_r)=\mathbb{I}\!\left[R_{\mathrm{pass}}(y_g)=1\right]\cdot\operatorname{clip}\!\left(1-\frac{t(y_g)}{t(y_r)},\,0,\,1\right),
\)
% \begin{equation}
% R_{\mathrm{eff}}(y_g,y_r)=
% \begin{cases}
% \displaystyle
% \operatorname{clip}\!\left(1-\frac{t(y_g)}{t(y_r)},\,0,\,1\right),
% &\begin{gathered}
% R_{\mathrm{pass}}(y_g)=1,
% % \text{ and valid timings are available},
% \end{gathered}\\[6pt]
% 0,&\text{otherwise},
% \end{cases}
% \label{eq:coding_efficiency}
% \end{equation}
where $t(y_g)$ is the execution time of the generated program and $t(y_r)$ is the mean execution time over four independent runs of the reference implementation. Both are measured using the same test harness.

\paragraph{Experimental setting.}
We use Qwen2.5-Coder-Instruct \citep{hui2024qwen25coder} at 0.5B, 1.5B, 3B, and 7B scales. Models are trained on the 2,641-problem training split of LeetCodeDataset \citep{xia2025leetcodedataset} and evaluated on its 228-problem test split. This dataset contains programming problems paired with reference solutions and executable tests. Training hyperparameters are provided in Appendix~\ref{app:training_coding}. To assess generalization without further training, we additionally evaluate HumanEval \citep{chen2021evaluating}, which tests function completion from natural-language specifications; MBPP \citep{austin2021program}, which contains basic Python programming tasks; and LiveCodeBench v6 \citep{jain2024livecodebench}, which contains competition programming problems. Table~\ref{tab:qwen25_combined_results} reports Executable, Pass@1, and Efficiency on LeetCodeDataset, together with Pass@1 on the three additional benchmarks. Executable measures the fraction of generated programs that execute without runtime errors, and Pass@1 measures the fraction that pass all tests. 
% Efficiency measures the percentage of eligible passing solutions that achieve a speedup of at least $1.01\times$ over the reference implementation.
Efficiency measures the percentage of eligible generated programs that run
strictly faster than their reference implementations, i.e.,
$\mathrm{speedup}=t_{\mathrm{ref}}/t_{\mathrm{gen}}>1$.
A pair is eligible when both programs pass all tests and have positive,
finite runtimes. We re-execute both programs in eight paired rounds
on the same CPU core, alternating their order, and report the mean
of the eight per-round percentages.
% Appendix~\ref{app:paired_runtime} details the timing protocol.

\paragraph{Main Results.}
\textbf{CorrGRPO consistently improves coding correctness across model scales and extends the empirical correctness--efficiency Pareto frontier.}
Table~\ref{tab:qwen25_combined_results} shows that CorrGRPO achieves the highest LeetCodeDataset Pass@1 and average benchmark Pass@1 at every evaluated model scale, outperforming both GRPO and GDPO. Relative to GRPO, average Pass@1 increases by 2.09, 0.80, 2.27, and 4.21 percentage points at 0.5B, 1.5B, 3B, and 7B, respectively. Figure~\ref{fig:pareto_frontier} further shows that CorrGRPO accounts for every nondominated configuration across the evaluated methods and model scales. Its 7B model achieves the highest Pass@1 at 24.12\%, its 0.5B model achieves the highest Efficiency at 67.86\%, and its 3B model occupies an intermediate frontier point with 14.47\% Pass@1 and 60.00\% Efficiency. 
% Furthermore, we demonstrate that CorrGRPO is compatible with GRPO variants, including GDPO~\cite{liu2026gdpo}, CISPO~\cite{minimax2025m1}, and DAPO~\cite{yu2025dapo}, with analysi in Appendix~\ref{app:corrgrpo_compatibility} and results in Table~\ref{tab:qwen25_combined_results}.

% Every evaluated base-model, GRPO, and GDPO configuration is dominated by at least one CorrGRPO configuration across these scales. 
% CorrGRPO therefore provides stronger observed operating points for different preferences between functional correctness and execution efficiency.

\subsection{Tool Calling}
\label{sec:exp_tools}

\paragraph{Task.}
We study structured tool calling, where a model selects the appropriate functions and generates their arguments from a user request and the available tool descriptions. Successful tool use requires jointly identifying the correct function, supplying the required parameter names, and assigning the correct parameter values. We evaluate both individual field correctness and complete-call correctness.

\input{latex/table-tool-call}

\paragraph{Reward design.}
Let $y_g$ denote the generated response and $y_r$ the reference response. We combine four reward components, suppressing their shared arguments $(y_g,y_r)$ for brevity:
\begin{equation}
R_{\mathrm{tool}}(y_g,y_r)=R_{\mathrm{fn}}+R_{\mathrm{pn}}+R_{\mathrm{pv}}+R_{\mathrm{fmt}}.
\label{eq:tool_reward}
\end{equation}
Here, $R_{\mathrm{fn}}$, $R_{\mathrm{pn}}$, and $R_{\mathrm{pv}}$ evaluate function names, parameter names, and parameter values, respectively, and $R_{\mathrm{fmt}}$ evaluates output format. Let $S_{\mathrm{fn}},S_{\mathrm{pn}},S_{\mathrm{pv}}\in[0,1]$ denote their matching scores. The reward components use the following scales:
\(R_{\mathrm{fn}}=0.5(2S_{\mathrm{fn}}-1),\ R_{\mathrm{pn}}=2S_{\mathrm{pn}}-1,\ R_{\mathrm{pv}}=1.5(2S_{\mathrm{pv}}-1).\)
% \begin{equation}
% R_{\mathrm{fn}}=0.5(2S_{\mathrm{fn}}-1),\qquad R_{\mathrm{pn}}=2S_{\mathrm{pn}}-1,\qquad R_{\mathrm{pv}}=1.5(2S_{\mathrm{pv}}-1).
% \label{eq:tool_reward_scales}
% \end{equation}
These rewards provide partial credit for incomplete calls. Parameter matching requires a matching function, and value correctness requires the corresponding parameter name. Appendix~\ref{app:tool_matching} provides the matching procedure and score definitions.
The format reward is 1 when the response follows the required structure and 0 otherwise. We provide the response format template in Appendix~\ref{app:tool-call-format}.

\paragraph{Experimental setting.}
We evaluate Qwen2.5-7B-Instruct \citep{yang2024qwen25}, Qwen3-4B-Thinking-2507 \citep{qwen2025thinking2507}, and Qwen3-8B \citep{yang2025qwen3}, comparing their base, GRPO, and CorrGRPO. We train on RLLA-4K from ToolRL \citep{qian2025toolrl}, which contains user requests, tool descriptions, and reference responses specifying tool calls or direct answers. Our split contains 3,920 training examples and 80 test examples. 
% tool-call metrics are computed on the 71 test examples with reference tool calls. 
% Training hyperparameters are provided in Appendix~\ref{app:training_tools}. 
To assess generalization without further training, we evaluate API-Bank \citep{li2023apibank}, a benchmark of tool-use dialogues with executable APIs. Table~\ref{tab:rlla-api-bank-accuracy} reports normalized function-name, parameter-name, and parameter-value scores on RLLA-4K, with the all-exact score, the percentage of examples for which all tool-call fields are correct. API-Bank results cover v1, v2, and v3, with final-call correctness assessed using its execution-based checker.

\input{latex/table-agent-security}

\paragraph{Main Results.}
\textbf{CorrGRPO improves complete-call correctness and generalization across all three evaluated backbones.} Table~\ref{tab:rlla-api-bank-accuracy} shows that CorrGRPO achieves the highest RLLA-4K and API-Bank average all-exact scores for every backbone. Compared with GRPO, RLLA-4K all-exact score increases from 63.38\% to 67.61\% for Qwen2.5-7B, from 57.75\% to 59.15\% for Qwen3-4B-Thinking, and from 61.97\% to 63.38\% for Qwen3-8B. These improvements accompany higher parameter-value scores across all three models, supporting more accurate argument generation. On API-Bank, the average score improves by 1.14, 3.94, and 2.26 percentage points, respectively. The largest gain occurs for Qwen3-4B-Thinking on API-Bank v3, where correctness increases from 52.24\% to 64.08\%. Together, these improvements raise the overall average by 2.68, 2.67, and 1.84 percentage points, demonstrating gains in both complete-call accuracy on the training-domain benchmark and transfer to API-Bank.

% Section 4.3 fragment: insert after Tool Calling in the manuscript.
% Requires amsmath, amssymb, and natbib.
% Merge section4_agent_references.bib into the manuscript bibliography.
% Existing Table 3 label: tab:qwen25-asb-opi-injecagent-valid.
% RL hyperparameters below describe saved defaults; per-checkpoint overrides
% have not been verified for every row of the results table.

% Section 4.3 fragment: insert after Tool Calling in the manuscript.
% Requires amsmath, amssymb, and natbib.
% Merge section4_agent_references_v2.bib into the manuscript bibliography.
% Existing Table 3 label: tab:qwen25-asb-opi-injecagent-valid.
% RL hyperparameters below describe saved defaults; per-checkpoint overrides
% have not been verified for every row of the results table.

\subsection{Agent Utility vs. Security}
\label{sec:exp_agents}

\paragraph{Task.}
We study tool-using agents that must complete legitimate user tasks while resisting indirect prompt injections embedded in tool outputs. An agent interacts with an environment through multiple tool calls, where malicious instructions may attempt to redirect its actions toward an attacker's objective. We evaluate both task completion and attack success to determine whether the agent remains useful under adversarial interference.

\paragraph{Reward design.}
For a valid attacked trajectory, let $y_g$ include the agent's responses, tool calls, and observations, with prompt-injection content inserted into the tool observations. We assign equal weights to utility and security:
\begin{equation}
R_{\mathrm{agent}}(y_g)=R_{\mathrm{util}}(y_g)+R_{\mathrm{sec}}(y_g).
\label{eq:agent_reward}
\end{equation}

The utility reward $R_{\mathrm{util}}$ is 1 if the agent successfully completes the user's task under attack and 0 otherwise. The security reward $R_{\mathrm{sec}}$ is 1 if the attacker's objective is not achieved and 0 otherwise. Both rewards are determined by the environment's task checkers. These objectives can compete: avoiding tool interactions may prevent an attack while also preventing completion of the user's task. Their combination rewards useful behavior and resistance to malicious instructions. 
% Appendix~\ref{app:agentdojo_protocol} provides the attack setup, an illustrative example, and the handling of invalid reward outcomes.

\paragraph{Experimental setting.}
We evaluate Qwen2.5-3B-Instruct, Qwen2.5-7B-Instruct \citep{yang2024qwen25}, and Qwen3-8B \citep{yang2025qwen3}, comparing their base, GRPO, and CorrGRPO variants. Training takes place in AgentDojo \citep{debenedetti2024agentdojo}. We construct a deterministic split grouped by user task, containing 1,584 training cases and 411 test cases; the latter comprise 21 clean cases and 390 attacked cases. 
% Training hyperparameters are provided in Appendix~\ref{app:training_agents}. 
To assess generalization without further training, we evaluate Agent Security Bench (ASB; \citealp{zhang2025asb}) using 400 paired clean and attacked cases per model, and InjecAgent \citep{zhan2024injecagent}. 
% Appendix~\ref{app:agent_protocols} describes the three benchmarks, the attacks used, and their tool-use and utility evaluation protocols. 
Table~\ref{tab:qwen25-asb-opi-injecagent-valid} reports clean utility, utility under attack, attack success rate (ASR), and Joint Accuracy for AgentDojo and ASB. For $N$ paired cases, Joint Accuracy is computed as
\(
% \mathrm{Joint\ Accuracy}=
\frac{1}{N}\sum_{i=1}^{N}(1-A_i)\frac{U_{\mathrm{clean},i}+U_{\mathrm{attack},i}}{2},\)
% \begin{equation}
% \mathrm{Joint\ Accuracy}
% =\frac{1}{N}\sum_{i=1}^{N}(1-A_i)\frac{U_{\mathrm{clean},i}+U_{\mathrm{attack},i}}{2},
% \label{eq:agent_joint_accuracy}
% \end{equation}
where $U_{\mathrm{clean},i}$ and $U_{\mathrm{attack},i}$ denote user-task success in the clean and attacked executions, and $A_i$ denotes attack success in the attacked execution.

\begin{figure}[t]
    \centering
    \includegraphics[width=0.95\linewidth]{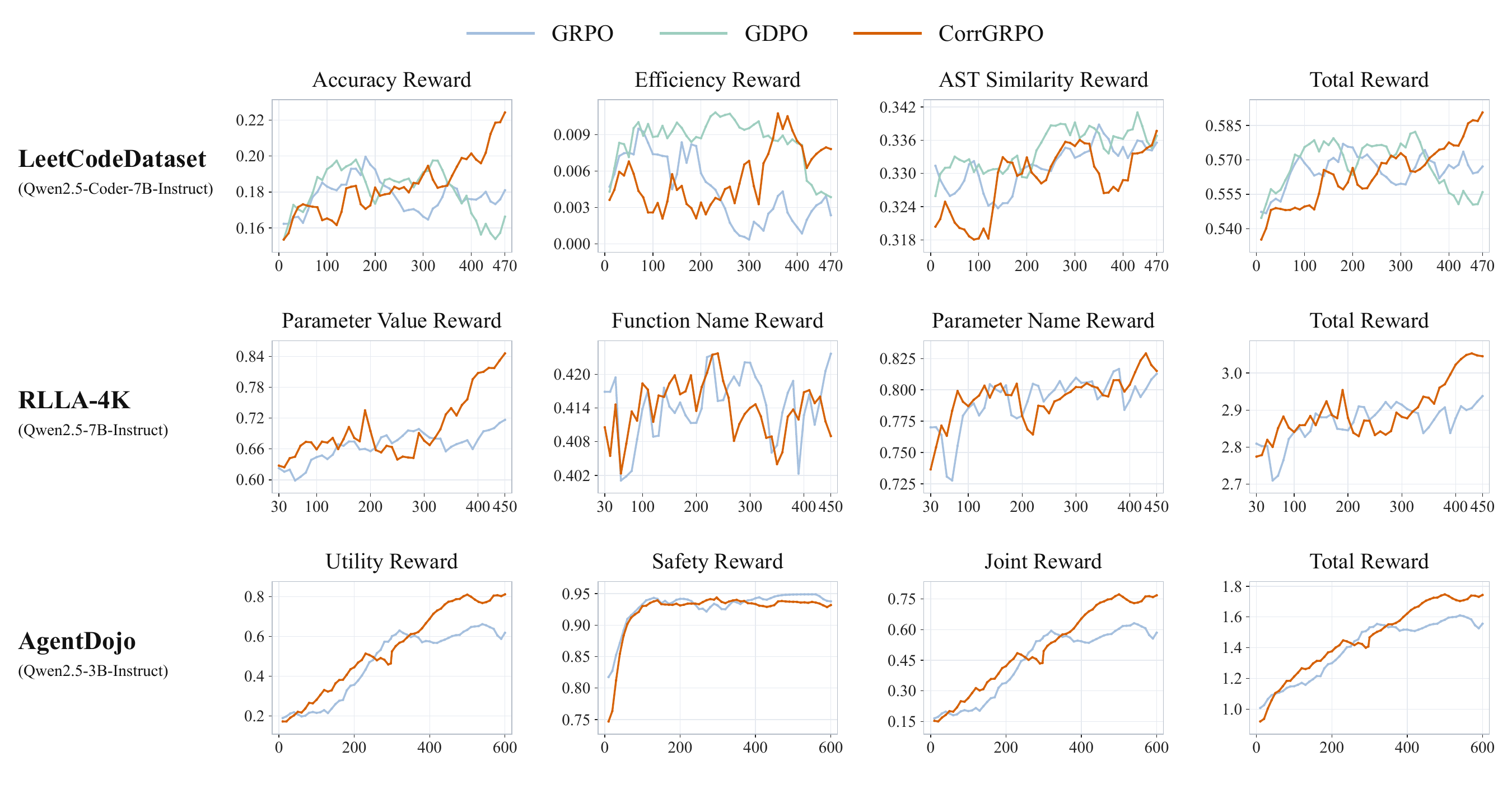}
    \caption{Validation reward dynamics of GRPO, GDPO, and CorrGRPO. Curves report validation mean@1 scores with exponential moving average smoothing (decay = 0.6).}
    \vspace{-0.1in}
    \label{fig:reward_curves}
    \vspace{-0.1in}
    
\end{figure}

\paragraph{Main Results.}
\textbf{CorrGRPO improves the balance between agent utility and security.} Relative to GRPO, CorrGRPO increases ASB Joint Accuracy from 14.88\% to 18.13\% for Qwen2.5-3B, from 31.38\% to 47.88\% for Qwen2.5-7B, and from 57.38\% to 58.63\% for Qwen3-8B (Table~\ref{tab:qwen25-asb-opi-injecagent-valid}). These gains accompany higher clean utility and utility under attack for all three models, supporting more effective task completion in adversarial environments. On InjecAgent, overall ASR decreases from 8.80\% to 5.52\%, from 13.53\% to 6.17\%, and from 13.94\% to 7.98\%, respectively, with reductions in every reported attack category. On AgentDojo, CorrGRPO also improves Joint Accuracy for Qwen2.5-3B and Qwen2.5-7B, with the largest gain at 3B, from 60.13\% to 89.62\%. 
% Together, the external benchmark results demonstrate improved joint performance on ASB and stronger resistance to prompt injection on InjecAgent.

%% file: latex/table-code-rl.tex
\begin{table*}[t]
\centering

\caption{
Coding RL results on LeetCodeDataset, HumanEval, MBPP, and LCB v6.
Efficiency is the mean percentage of eligible programs that run faster than the reference code.
% Efficiency denotes the percentage of eligible passing samples with
% $\mathrm{speedup} > 1.0$, averaged over eight paired execution rounds.
% Reference and generated programs are timed sequentially on the same CPU core,
% with alternating execution order.
% A pair is eligible when both programs pass all tests and have valid runtimes.
Executable denotes running time success rate.
Avg.\ is the arithmetic mean of the Pass@1 evaluation results across 4 datasets.
% The best and second-best results within each model-scale group are shown in bold and underlined, respectively.
}
\vspace{-0.1in}

\label{tab:qwen25_combined_results}

\begingroup
\codertablefontsize
\setlength{\tabcolsep}{3pt}
\begin{tabular*}{\textwidth}{@{\extracolsep{\fill}}lccc@{\hspace{1pt}}cccc@{}}
\toprule
\multirow{2}{*}{Model}
& \multicolumn{3}{c}{LeetCodeDataset (In-dataset Eval)}
& HumanEval & MBPP & LCB v6 & Avg. \\
\cmidrule(lr){2-4}
\cmidrule(lr){5-5}
\cmidrule(lr){6-6}
\cmidrule(lr){7-7}
\cmidrule(lr){8-8}
& Efficiency & Executable & Pass@1
& Pass@1 & Pass@1 & Pass@1 & Pass@1 \\
\midrule

Qwen2.5-Coder-0.5B-Instruct & 50.00 & 61.40 & 1.75 & 54.88 & 33.60 & \underline{5.14} & 23.84 \\
\quad + GRPO & 60.71 & 87.72 & 3.07 & \underline{57.32} & \underline{40.00} & 3.43 & \underline{25.96} \\
\quad + GDPO & \underline{60.94} & \underline{88.60} & \underline{3.51} & 57.32 & 38.40 & 4.57 & 25.95 \\
\quad + CorrGRPO & \textbf{67.86} & \textbf{89.04} & \textbf{3.95} & \textbf{59.76} & \textbf{42.20} & \textbf{6.29} & \textbf{28.05} \\

\midrule

Qwen2.5-Coder-1.5B-Instruct & \textbf{58.93} & 75.44 & 3.07 & 61.59 & 54.40 & 10.86 & 32.48 \\
\quad + GRPO & 40.00 & 84.21 & 6.58 & \underline{64.63} & \underline{58.00} & \textbf{13.14} & \underline{35.59} \\
\quad + GDPO & 45.00 & \underline{85.09} & \underline{7.02} & 63.41 & \textbf{58.60} & \underline{12.57} & 35.40 \\
\quad + CorrGRPO & \underline{46.71} & \textbf{86.84} & \textbf{8.77} & \textbf{68.90} & 57.60 & 10.29 & \textbf{36.39} \\

\midrule

Qwen2.5-Coder-3B-Instruct & 54.41 & 77.63 & 7.46 & \underline{82.32} & 60.80 & \underline{17.14} & 41.93 \\
\quad + GRPO & \underline{56.00} & \textbf{83.77} & \underline{12.28} & 80.49 & 63.80 & 16.57 & 43.29 \\
\quad + GDPO & 48.96 & \underline{83.77} & 10.96 & 81.71 & \underline{64.40} & 16.57 & \underline{43.41} \\
\quad + CorrGRPO & \textbf{60.00} & 81.58 & \textbf{14.47} & \textbf{82.32} & \textbf{66.00} & \textbf{19.43} & \textbf{45.56} \\

\midrule

Qwen2.5-Coder-7B-Instruct & 44.35 & 80.70 & 14.91 & \underline{78.66} & 76.40 & 20.57 & 47.64 \\
\quad + GRPO & \underline{44.70} & \textbf{85.53} & 15.79 & 73.78 & 78.40 & 21.14 & 47.28 \\
\quad + GDPO & 41.28 & \underline{84.21} & \underline{16.23} & 76.83 & \textbf{79.60} & \underline{22.86} & \underline{48.88} \\
\quad + CorrGRPO & \textbf{51.42} & 80.70 & \textbf{24.12} & \textbf{78.66} & \underline{78.60} & \textbf{24.57} & \textbf{51.49} \\

\bottomrule
\end{tabular*}
\par
\endgroup

\vspace{-0.1in}

\end{table*}

%% file: latex/table-tool-call.tex
\begin{table*}[t]
\centering
\caption{Tool-call results on RLLA-4K and API-Bank.
API-Bank scores use the same all-exact metric for the correctness of final tool call.
% API-Bank Avg. is the mean across v1, v2, and v3.
The rightmost Avg. is the arithmetic mean of RLLA-4K all-exact score
and API-Bank Avg. score.
% Within each backbone group and metric, the best result is shown in bold and the second-best is underlined.
}
\label{tab:rlla-api-bank-accuracy}
\vspace{-0.1in}

\begingroup
\toolcalltablefontsize
\setlength{\tabcolsep}{3pt}
% \begin{tabular*}{\textwidth}{@{\extracolsep{\fill}}lccccccccc@{}}
\resizebox{\textwidth}{!}{%
\begin{tabular}{@{}lccccccccc@{}}
\toprule
\multirow{2}{*}{Model}
& \multicolumn{4}{c}{RLLA-4K (In-dataset Eval) }
& \multicolumn{4}{c}{API-Bank}
& \multirow{2}{*}{Avg.} \\
\cmidrule(lr){2-5}
\cmidrule(lr){6-9}
& Function & Param. Name & Param. Value & All Exact
& V1 & V2 & V3 & Avg. & \\
\midrule

Qwen2.5-7B-Instruct
  & 73.17 & 69.77 & 54.78 & 38.03
  & 17.29 & 19.26 & 24.49 & 20.35 & 29.19 \\
\quad + GRPO
  & \textbf{97.65} & \textbf{95.60}
  & \underline{76.91} & \underline{63.38}
  & \underline{79.45} & \underline{44.44}
  & \textbf{35.92} & \underline{53.27}
  & \underline{58.33} \\
\quad + CorrGRPO
  & \underline{95.07} & \underline{94.95}
  & \textbf{82.14} & \textbf{67.61}
  & \textbf{81.45} & \textbf{46.67}
  & \underline{35.10} & \textbf{54.41}
  & \textbf{61.01} \\

\midrule

Qwen3-4B-Thinking-2507
  & 33.80 & 31.46 & 25.35 & 21.13
  & 53.38 & 39.26 & 31.43 & 41.36 & 31.25 \\
\quad + GRPO
  & \textbf{94.84} & \underline{92.08}
  & \underline{72.07} & \underline{57.75}
  & \textbf{79.20} & \underline{52.59}
  & \underline{52.24} & \underline{61.34}
  & \underline{59.55} \\
\quad + CorrGRPO
  & \underline{93.78} & \textbf{92.31}
  & \textbf{73.74} & \textbf{59.15}
  & \underline{78.45} & \textbf{53.33}
  & \textbf{64.08} & \textbf{65.29}
  & \textbf{62.22} \\

\midrule

Qwen3-8B
  & 86.85 & 84.80 & 67.50 & 50.70
  & 74.69 & 49.63 & \textbf{46.53}
  & 56.95 & 53.83 \\
\quad + GRPO
  & \underline{92.72} & \underline{91.43}
  & \underline{72.54} & \underline{61.97}
  & \underline{78.20} & \underline{51.85}
  & 40.82 & \underline{56.95}
  & \underline{59.46} \\
\quad + CorrGRPO
  & \textbf{97.65} & \textbf{96.60}
  & \textbf{76.44} & \textbf{63.38}
  & \textbf{81.70} & \textbf{51.85}
  & \underline{44.08} & \textbf{59.21}
  & \textbf{61.30} \\

\bottomrule
% \end{tabular*}
\end{tabular}%
}
\par
\endgroup
\vspace{-0.15in}

\end{table*}

%% file: latex/table-agent-security.tex
\begin{table*}[t]
\centering

\caption{
% Agent utility-security results on AgentDojo, Agent Security Bench, and InjecAgent. 
% Utility represents agent tool calling success rate, while ASR denotes injection attack success rate. 
% Joint accuracy consider both, which is calculated by (1-ASR) * clean utility + utility under attack.
Agent utility and security results. Utility measures the agent’s tool-calling success rate, while ASR measures the success rate of prompt injection attacks. Joint accuracy combines utility and security as $(1-\mathrm{attack\_success}) \times \frac{{\mathrm{clean\_utility}} + {\mathrm{utility\_under\_attack}}}{2}$. In InjecAgent, all reported scores represent ASR.
% Clean utility and utility under attack are original-task success rates without attack and under injection attack, respectively.
% All values are percentages. 
% Paired entries report AgentDojo / ASB OPI, respectively. 
% $\uparrow$ indicates higher is better; $\downarrow$ indicates lower is better.
}
\label{tab:qwen25-asb-opi-injecagent-valid}
\vspace{-0.1in}

\begingroup
\securitytablefontsize
\setlength{\tabcolsep}{5pt}
% \renewcommand{\arraystretch}{1.12}
% Account for the cmidrule spacing in the InjecAgent header.
\newcommand{\qwenheaderoffset}{%
\dimexpr-\aboverulesep/2-\belowrulesep/2-\cmidrulewidth/2\relax}
\begin{tabular*}{\textwidth}{@{\extracolsep{\fill}}lcccc@{}}
\toprule
\multicolumn{5}{@{}l}{\textbf{AgentDojo (In-dataset Eval) / Agent Security Bench}} \\
\midrule
Model & Clean Utility $\uparrow$ & Utility Under Attack $\uparrow$
& ASR $\downarrow$ & Joint Accuracy $\uparrow$ \\
\midrule
Qwen2.5-3B-Instruct & 28.57 / 10.00 & 14.62 / 6.25 & 3.85 / \textbf{8.00} & 18.97 / 6.13 \\
\hspace{0.35em}+ GRPO & \underline{52.38} / \underline{21.75} & \underline{66.92} / \underline{9.75} & \textbf{0.00} / \underline{9.00} & \underline{60.13} / \underline{14.88} \\
\hspace{0.35em}+ CorrGRPO & \textbf{95.24} / \textbf{26.00} & \textbf{84.36} / \textbf{13.25} & \underline{1.54} / 9.25 & \textbf{89.62} / \textbf{18.13} \\
\midrule
Qwen2.5-7B-Instruct & 38.10 / \textbf{64.75} & 30.51 / \underline{47.00} & 11.28 / 33.25 & 28.72 / \underline{42.00} \\
\hspace{0.35em}+ GRPO & \textbf{90.48} / 30.00 & \underline{80.00} / 41.75 & \textbf{0.26} / \textbf{26.50} & \underline{84.49} / 31.38 \\
\hspace{0.35em}+ CorrGRPO & \underline{85.71} / \underline{59.75} & \textbf{83.08} / \textbf{53.25} & \underline{1.28} / \underline{29.00} & \textbf{86.03} / \textbf{47.88} \\
\midrule
Qwen3-8B & 71.43 / 60.00 & 32.31 / 51.50 & 19.49 / 31.25 & 40.38 / 49.63 \\
\hspace{0.35em}+ GRPO & \textbf{85.71} / \underline{60.00} & \textbf{77.18} / \underline{61.50} & \underline{2.05} / \textbf{19.50} & \textbf{80.26} / \underline{57.38} \\
\hspace{0.35em}+ CorrGRPO & \underline{85.71} / \textbf{70.00} & \underline{75.13} / \textbf{61.75} & \textbf{1.54} / \underline{22.50} & \underline{79.62} / \textbf{58.63} \\
% \bottomrule
\end{tabular*}
% \par\vspace{8pt}
\begin{tabular*}{\textwidth}{@{\extracolsep{\fill}}lccccc@{}}
\toprule
\multicolumn{6}{@{}l}{\textbf{InjecAgent}} \\
\midrule
\multirow[c]{2}{*}[\qwenheaderoffset]{Model}
& \multirow[c]{2}{*}[\qwenheaderoffset]{Direct Harm $\downarrow$}
& \multicolumn{3}{c}{Data Stealing}
& \multirow[c]{2}{*}[\qwenheaderoffset]{Total $\downarrow$} \\
\cmidrule(lr){3-5}
& & S1 $\downarrow$ & S2 $\downarrow$ & Total $\downarrow$ & \\
\midrule
Qwen2.5-3B-Instruct & 18.38 & 50.29 & \underline{76.39} & 34.59 & 27.12 \\
\hspace{0.35em}+ GRPO & \underline{5.03} & \underline{20.11} & 79.31 & \underline{13.07} & \underline{8.80} \\
\hspace{0.35em}+ CorrGRPO & \textbf{4.67} & \textbf{17.26} & \textbf{52.63} & \textbf{6.33} & \textbf{5.52} \\
\midrule
Qwen2.5-7B-Instruct & 33.85 & 44.21 & 92.24 & 38.91 & 36.17 \\
\hspace{0.35em}+ GRPO & \underline{16.49} & \underline{15.06} & \underline{75.00} & \underline{10.12} & \underline{13.53} \\
\hspace{0.35em}+ CorrGRPO & \textbf{6.91} & \textbf{9.77} & \textbf{60.00} & \textbf{5.44} & \textbf{6.17} \\
\midrule
Qwen3-8B & 30.56 & 42.22 & 92.13 & 38.79 & 34.86 \\
\hspace{0.35em}+ GRPO & \underline{13.45} & \underline{16.36} & \underline{88.75} & \underline{14.37} & \underline{13.94} \\
\hspace{0.35em}+ CorrGRPO & \textbf{6.94} & \textbf{12.02} & \textbf{74.19} & \textbf{8.91} & \textbf{7.98} \\
\bottomrule
\end{tabular*}

\par
\vspace{2pt}
\endgroup
\vspace{-0.15in}

\end{table*}

%% file: latex/5-related-works.tex
\section{Related Work}

Multi-reward RL algorithms differ primarily in how they aggregate reward signals and balance their contributions during policy optimization. Scalarization-based approaches combine multiple rewards into a single training signal: MORLAIF trains separate preference models for individual objectives and aggregates their scores through scalarization functions before PPO updates \citep{williams2024multiobjective}. To adapt objective tradeoffs during training, Safe RLHF maximizes helpfulness subject to safety constraints using dynamically updated Lagrange multipliers \citep{dai2023saferlhf}, while dynamic reward weighting adjusts scalarization weights through hypervolume-guided adaptation or gradient-based optimization \citep{lu2025dynamicweighting}. More closely related to our work, recent methods modify how multiple rewards enter group-relative advantage estimation. MO-GRPO automatically reweights reward functions according to their variances to balance their contributions \citep{ichihara2025mogrpo}, whereas GDPO independently normalizes each reward within rollout groups, aggregates the resulting advantages, and applies batch-level normalization to stabilize their overall magnitude \citep{liu2026gdpo}. RDPO further addresses reward dependence by combining magnitude-aware quantile normalization with Mahalanobis whitening before aggregation, reducing redundant variation among correlated reward dimensions \citep{bai2026rdpo}. Our CorrGRPO instead retains the centered total reward and its prescribed relative weights, while replacing the covariance-based normalization denominator with an aggregate of Pearson correlations. 
% This design makes advantage scaling responsive to inter-reward dependence while removing the multiplicative influence of component scales from the normalization denominator.

%% file: latex/6-conclusion.tex
\section{Conclusion}
We introduced CorrGRPO, a correlation-normalized variant of GRPO for multi-reward reinforcement learning. Our analysis shows that normalizing the summed reward by its within-group standard deviation implicitly aggregates all pairwise reward covariances, coupling reward dependence with component scales. CorrGRPO replaces these covariance terms with Pearson correlation coefficients while preserving the centered total reward and its prescribed relative weights. This modification allows advantage normalization to respond to inter-reward correlations without being dominated by reward components with larger within-group standard deviations. Experiments on code generation, tool calling, and agent security, using models ranging from 0.5B to 8B parameters, demonstrate improved performance and an outward expansion of the empirical Pareto frontier between competing objectives. These results highlight the value of explicitly accounting for inter-reward correlations when designing advantage estimators for multi-reward learning.

\section*{AI Use Statement}

We used generative AI tools to improve the clarity of expression, language quality, and grammatical correctness of the manuscript. Beyond the language models studied in our experiments, we did not use generative AI tools to develop the proposed method, formulate mathematical claims, write proofs, design experiments, implement methods, generate or process experimental datasets, or interpret results. The authors reviewed and revised all AI-assisted text to ensure accuracy and consistency with the underlying research. We take full responsibility for the final manuscript, including all claims, results, and AI-assisted content.

\section*{Ethics Statement}

This work studies multi-reward optimization for language models through existing benchmarks for code generation, tool calling, and agent security. Improving these capabilities may benefit legitimate applications but may also facilitate harmful automation or unauthorized tool use. Our security experiments evaluate resistance to benchmark prompt-injection attacks; improvements on these benchmarks do not establish safety in real-world deployments or against unseen attacks. Moreover, correlation-based normalization does not correct biased or misspecified rewards, and the resulting behavior remains dependent on the selected objectives and their weights. Deployment therefore requires application-specific safety evaluation, appropriate tool-access restrictions, and human oversight.

\section*{Reproducibility Statement}

Section~\ref{sec:corrgrpo} defines CorrGRPO, and Appendix~\ref{sec:corrgrpo-implementation} provides its core implementation, including numerical stabilization and the handling of zero-variance reward components. Sections~\ref{sec:coding_reasoning}--\ref{sec:exp_agents} describe the model backbones, reward functions, experimental settings, and evaluation metrics. Appendix~\ref{app:training_settings} reports training hyperparameters, Appendix~\ref{app:reward_details} details reward computation, and Appendix~\ref{sec:datasets} describes the datasets, evaluation subsets, and benchmark settings. Representative prompts and responses are provided in Appendix~\ref{app:training_examples}. Appendix~\ref{app:corrgrpo_analysis} presents the assumptions and derivations underlying the theoretical properties, while Appendix~\ref{app:compatibility} details compatibility with other optimization methods. These materials support reproduction and inspection of the proposed method and its evaluation.

%% file: latex/5-discussion.tex
\section{Further Discussion} 
\noindent\textbf{Preserving groupwise gradient directions.}
Let $L_{\mathrm{GRPO},q}$ and $L_{\mathrm{CorrGRPO},q}$ denote the clipped
reward surrogates averaged over tokens and trajectories in a fixed rollout
group for prompt $q$, excluding the KL term. Let
$\mathbf C=[\hat\rho_{lm}]$ be the group's sample correlation matrix,
$\mathbf s=(\hat\sigma_1,\ldots,\hat\sigma_r)^\top$ its vector of reward
standard deviations, and $\mathbf1$ the all-ones vector. As derived in
Appendix~\ref{app:corrgrpo_rescaling},
Equation~\ref{eq:app_corr_rescaling}, the shared numerator gives
\begin{equation}
A_{\mathrm{CorrGRPO}}^i=c_qA_{\mathrm{GRPO}}^i,
\qquad
c_q=\frac{\sqrt{\mathbf s^\top\mathbf C\mathbf s}+\varepsilon}
{\sqrt{\mathbf1^\top\mathbf C\mathbf1}+\varepsilon}>0.
\label{eq:corrgrpo_group_coefficient}
\end{equation}
The coefficient $c_q$ is computed from the same group's reward statistics,
shared across its trajectories, and held fixed during surrogate optimization.
Because the clipped surrogate is positively homogeneous in the advantage,
this scaling preserves the active clipping branches and yields
\begin{equation}
\nabla_\theta L_{\mathrm{CorrGRPO},q}
=c_q\nabla_\theta L_{\mathrm{GRPO},q},
\qquad c_q>0.
\label{eq:corrgrpo_group_gradient}
\end{equation}
The derivation is given in
Appendix~\ref{app:corrgrpo_rescaling},
Equations~\ref{eq:app_corr_group_surrogate}--\ref{eq:app_corr_group_gradient}.
Thus, at the same policy parameters, CorrGRPO preserves the direction of each
group's nonzero reward-driven gradient while adapting its magnitude to reward
dependence. The clipping thresholds remain unchanged. Since $c_q$ can differ
across groups, their relative contributions to the batch gradient can change;
the KL term retains its original coefficient.

\noindent\textbf{A quadratic-norm view of reward fluctuations.}
Using the same correlation matrix $\mathbf C$, define
$\|\mathbf x\|_{\mathbf C}=\sqrt{\mathbf x^\top\mathbf C\mathbf x}$,
which is a norm when $\mathbf C$ is positive definite and a seminorm
when it is singular. 
% As derived in Appendix~\ref{app:corrgrpo_quadratic},
% Equations~\eqref{eq:app_corr_shared_metric} and~\eqref{eq:app_corr_induced_norm},
The two normalization statistics use the same correlation metric with
different input vectors:
\begin{equation}
\begin{aligned}
\text{GRPO:}\quad
\widehat{\mathrm{Var}}(R)
&=\mathbf s^\top\mathbf C\mathbf s
 =\|\mathbf s\|_{\mathbf C}^2,\\
\text{CorrGRPO:}\quad
\sum_{l,m}\hat\rho_{lm}
&=\mathbf1^\top\mathbf C\mathbf1
 =\|\mathbf1\|_{\mathbf C}^2.
\end{aligned}
\label{eq:corrgrpo_shared_metric}
\end{equation}
Their advantage denominators are consequently
$\|\mathbf s\|_{\mathbf C}+\varepsilon$ and
$\|\mathbf1\|_{\mathbf C}+\varepsilon$, respectively.

This admits a signal-processing interpretation. Treat the standardized,
centered reward components as correlated input channels to a linear combiner.
For a channel-gain vector $\mathbf x$, the output's sample variance is
$\mathbf x^\top\mathbf C\mathbf x$.
% (Appendix~\ref{app:corrgrpo_quadratic},
% Equation~\eqref{eq:app_corr_linear_combiner}). 
Diagonal terms measure individual
channel contributions, while off-diagonal terms capture reinforcement or
cancellation between channels. GRPO uses $\mathbf x=\mathbf s$, restoring each source's
own fluctuation amplitude before measuring the combined output. Its
normalization therefore depends on both the correlation structure and the
individual source amplitudes. CorrGRPO uses $\mathbf x=\mathbf1$, giving all
standardized channels equal gain so that the combined fluctuation reflects
their correlations without additional weighting by their original scales.

% In Figure~\ref{fig:example}, $\mathbf s$ is proportional to $(1,1,9)^\top$, so GRPO weights
% the third channel's diagonal contribution $81$ times as much as either of the
% first two and its cross terms $9$ times as much as the first pair's, before
% accounting for their correlations. CorrGRPO replaces these unequal gains with
% $(1,1,1)^\top$, allowing the strong correlation between $r_1$ and $r_2$ to
% play a correspondingly larger role. This equal-gain construction applies to
% the normalization statistic; the numerator continues to aggregate the
% original rewards with their specified weights.

\noindent\textbf{Compatibility with other policy optimization methods.}
CorrGRPO is compatible with other RL policy optimization methods, including
DAPO~\citep{yu2025dapo}, CISPO~\citep{minimax2025m1}, and
GDPO~\citep{liu2026gdpo}. For DAPO and CISPO, the original group-relative
advantage can be directly replaced with $A_{\mathrm{CorrGRPO}}^i$, while
retaining their respective policy objectives and clipping operations. For
GDPO, we retain its per-reward group normalization and aggregation, and
replace only the final batch standard-deviation denominator with a
correlation norm computed across the batch from the resulting per-reward
advantage components. The corresponding objectives are derived in
Appendices~\ref{app:corrgrpo_dapo}, \ref{app:corrgrpo_cispo}, and
\ref{app:corrgrpo_gdpo}, respectively.
We also evaluate this integration with CISPO on Qwen2.5-Coder-7B-Instruct,
with results reported in Appendix~\ref{sec:corrgrpo_compatibility_experiments}.

% \paragraph{Slower entropy decay.} Figure~12 shows that CorrGRPO preserves higher policy entropy over substantial portions of training, with slower entropy decay particularly evident on AgentDojo and Qwen2.5-Coder-7B-Instruct on LeetCodeDataset. Although this pattern varies across settings, these results suggest that CorrGRPO can sustain policy entropy for longer during optimization. One hypothesis is that its correlation-based normalization adjusts the magnitude of reward-driven updates according to reward correlations, which may moderate premature concentration on a narrow set of responses. The retained entropy is a key resource for continued performance improvement: it preserves opportunities to explore alternative reasoning paths and discover higher-reward responses after the policy has learned initially successful behaviors. 
% % By contrast, an early loss of diversity can restrict further exploration and make subsequent gains harder to achieve. 
% CorrGRPO's more sustained entropy therefore offers a plausible explanation for its ability to support continued improvements, with the benefit arising from maintaining useful exploration.

\paragraph{Slower entropy decay.} Figure~\ref{fig:training-diagnostics} shows that CorrGRPO preserves higher policy entropy over training, with slower entropy decay particularly evident on AgentDojo and LeetCodeDataset. Although this pattern varies across settings, these results suggest that CorrGRPO can sustain policy entropy for longer during optimization. As shown by \citet{cui2025entropy}, entropy collapse diminishes exploration and accompanies performance saturation, while entropy control enables continued exploration and improves downstream performance. From this perspective, the higher entropy retained by CorrGRPO helps explain its stronger performance: maintaining policy diversity preserves opportunities to explore alternative reasoning paths and discover higher-reward responses, supporting continued improvement beyond initially successful behaviors.

%% file: latex/7-appendix.tex
\input{latex/appendix-core-implementation}

\input{latex/appendix-runtime_pass_correlation}

\input{latex/appendix-all-training-curves}
\input{latex/appendix-original-scores}

\input{latex/appendix-compatibility}

\input{latex/appendix-rl-background}

\input{latex/appendix-data-examples}
\input{latex/appendix-train-setting}

\input{latex/appendix-reward-computation}

\input{latex/appendix-adv-vs-std-corr}

\input{latex/appendix-equation}

\input{latex/appendix-dataset-intro}

% \input{latex/appendix-agent-security}

%% file: latex/appendix-core-implementation.tex
\section{Core Implementation of CorrGRPO}
\label{sec:corrgrpo-implementation}

We compare the advantage implementation for GRPO and CorrGRPO in Listing~\ref{lst:corrgrpo-diff}.
For each prompt group, \texttt{group\_scores} contains the $n\times r$
reward components with their configured weights already applied.
% CorrGRPO retains GRPO's centered total reward and replaces its
% standard-deviation denominator with the square root of the sum of
% pairwise Pearson correlations. 
The implementation uses sample covariance
with divisor $n-1$, assigns zero rows and columns to zero-variance
components, and clamps the correlation sum to zero to guard against
negative round-off. 
% The excerpt shows the normalized branch for $n>1$;
% the full CorrGRPO function returns zero advantages for singleton groups
% and skips division when normalization is disabled. Token broadcasting
% and masking are unchanged. 
The listing follows the source's numerical convention:
CorrGRPO uses $\sqrt{\max(\sum_{l,m}\hat\rho_{lm},0)+\epsilon}$
with $\epsilon=10^{-6}$, whereas the paper's formula and the GRPO
implementation place $\epsilon$ outside the square root.

\begingroup
\definecolor{corrgrpoadd}{RGB}{0,105,55}
\definecolor{corrgrpodel}{RGB}{160,35,35}
\lstset{
  basicstyle=\ttfamily\footnotesize,
  columns=fullflexible,
  keepspaces=true,
  showstringspaces=false,
  breaklines=true,
  frame=single,
  rulecolor=\color{black!25},
  framerule=0.4pt,
  xleftmargin=0.5em,
  xrightmargin=0.5em,
  framexleftmargin=0.4em,
  framexrightmargin=0.4em,
  captionpos=t,
  aboveskip=1em,
  belowskip=0.5em,
  morecomment=[f][\color{corrgrpoadd}]{+},
  morecomment=[f][\color{corrgrpodel}]{-}
}
\begin{lstlisting}[caption={GRPO to CorrGRPO: abbreviated groupwise diff. Red lines are removed and green lines added; reward loading and group indexing are omitted.},label={lst:corrgrpo-diff}]
--- GRPO
+++ CorrGRPO
 # group_scores: [n, r], weighted rewards; n > 1
 # Executed within torch.no_grad().
 centered_scores = group_scores - group_scores.mean(0, keepdim=True)
 centered_total_reward = centered_scores.sum(dim=-1)
-denominator = group_scores.sum(dim=-1).std() + epsilon
+covariance = centered_scores.T @ centered_scores / (group_scores.size(0) - 1)
+variances = covariance.diagonal()
+inverse_std = torch.where(
+    variances > 0,
+    torch.rsqrt(variances),
+    torch.zeros_like(variances),
+)
+covariance_coefficients = (
+    covariance * inverse_std[:, None] * inverse_std[None, :]
+)
+coefficient_sum = covariance_coefficients.sum().clamp_min(0)
+denominator = torch.sqrt(coefficient_sum + epsilon)
 group_advantages = centered_total_reward / denominator
 scalar_advantages.index_copy_(0, positions_tensor, group_advantages)

 advantages = scalar_advantages.unsqueeze(-1) * response_mask
 return advantages, advantages
\end{lstlisting}
\endgroup

%% file: latex/appendix-runtime_pass_correlation.tex
\section{Correlation Analysis for Two Dependent Rewards}

This section examines how two dependent rewards jointly shape CorrGRPO's
advantage normalization as training progresses. We use runtime success and
functional correctness in coding RL as an example, deriving their correlation
and examining training dynamics on LeetCodeDataset.

\paragraph{Correlation analysis.}
Let $R_{\mathrm{run}}\in\{0,1\}$ indicate whether a generated program executes
without runtime errors, and $R_{\mathrm{pass}}\in\{0,1\}$ indicate whether it
passes all test cases. Since passing all tests requires successful execution,
$R_{\mathrm{run}}^iR_{\mathrm{pass}}^i=R_{\mathrm{pass}}^i$.
Denoting their within-group means by $\bar R_{\mathrm{run}}$ and
$\bar R_{\mathrm{pass}}$, their Pearson correlation, when both variances are
nonzero, satisfies
\begingroup
\small
\begin{equation}
\widehat\rho_{\mathrm{run},\mathrm{pass}}
=\frac{\widehat{\operatorname{Cov}}(R_{\mathrm{run}},R_{\mathrm{pass}})}
{\sqrt{\widehat{\operatorname{Var}}(R_{\mathrm{run}})
\widehat{\operatorname{Var}}(R_{\mathrm{pass}})}}
=\frac{\bar R_{\mathrm{pass}}-\bar R_{\mathrm{run}}\bar R_{\mathrm{pass}}}
{\sqrt{\bar R_{\mathrm{run}}(1-\bar R_{\mathrm{run}})
\bar R_{\mathrm{pass}}(1-\bar R_{\mathrm{pass}})}}
=\sqrt{\frac{\bar R_{\mathrm{pass}}(1-\bar R_{\mathrm{run}})}
{\bar R_{\mathrm{run}}(1-\bar R_{\mathrm{pass}})}}.
\label{eq:runtime-pass-correlation}
\end{equation}
\endgroup
Once runtime success stabilizes at a fixed level below one, increasing the
mean pass reward $\bar R_{\mathrm{pass}}$ increases the correlation
$\widehat\rho_{\mathrm{run},\mathrm{pass}}$.
With equal reward weights, CorrGRPO computes
\begin{equation}
A^i_{\mathrm{CorrGRPO}}
=\frac{(R^i_{\mathrm{run}}-\bar R_{\mathrm{run}})
+(R^i_{\mathrm{pass}}-\bar R_{\mathrm{pass}})}
{\sqrt{2+2\widehat\rho_{\mathrm{run},\mathrm{pass}}}+\varepsilon}.
\label{eq:runtime-pass-advantage}
\end{equation}
As the two rewards become increasingly correlated, CorrGRPO enlarges the
denominator and attenuates their combined signal for a fixed centered total
reward, moderating the effect of redundant reward information on the policy
update. In the limiting case $R_{\mathrm{run}}=R_{\mathrm{pass}}$, ignoring
$\varepsilon$ gives
$A^i_{\mathrm{CorrGRPO}}=R^i_{\mathrm{run}}-\bar R_{\mathrm{run}}$:
duplicating the same reward does not double the advantage.

\begin{figure}[htbp]
\centering
\includegraphics[width=0.98\linewidth]{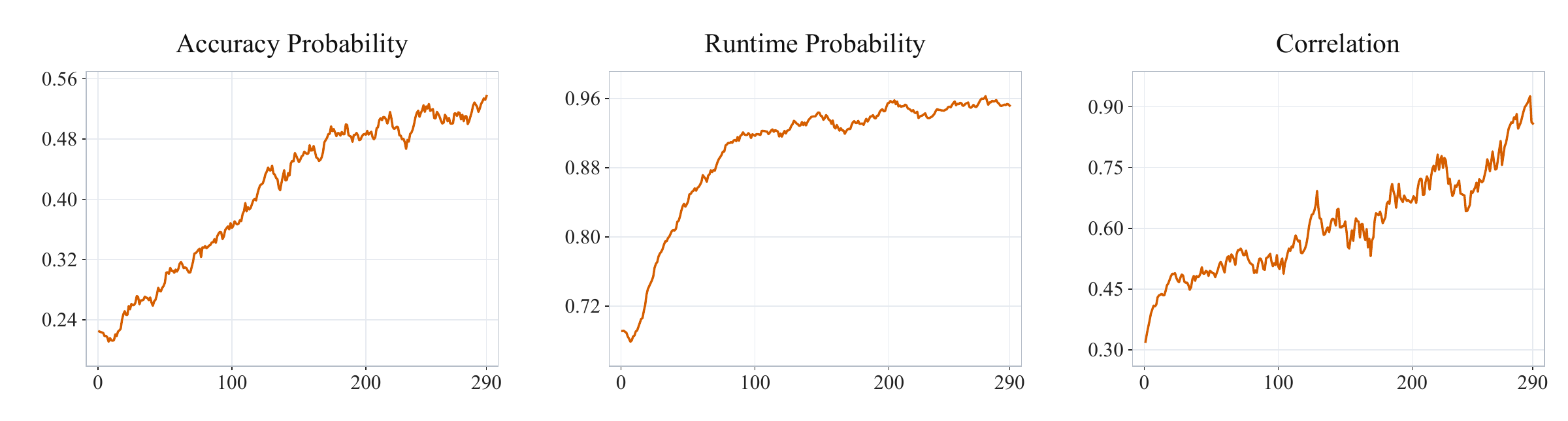}
\caption{Training dynamics on LeetCodeDataset using equally weighted
runtime-success and passing rewards.}
\label{fig:runtime-pass-curves}
\end{figure}

\paragraph{Experimental analysis.}
We train on LeetCodeDataset using only runtime-success and all-test passing
rewards with equal weights,
$R_{\mathrm{code}}=R_{\mathrm{run}}+R_{\mathrm{pass}}$.
As shown in Figure~\ref{fig:runtime-pass-curves}, runtime success improves
rapidly and approximately converges first, remaining around $0.92$--$0.96$
after about $100$ steps. During this plateau, the pass rate continues to rise
from approximately $0.36$ to $0.54$, while the reported reward correlation
also increases overall. This trend qualitatively agrees with the analysis:
as more executable programs pass all tests, the two reward signals increasingly
overlap. Under CorrGRPO's normalization, stronger correlation increases the
denominator and reduces the shared advantage scale, moderating the
reinforcement from these overlapping signals while preserving their equal
weights.

%% file: latex/appendix-all-training-curves.tex
% Requires \usepackage{graphicx} in the main preamble.

\section{Additional Training Dynamics}
\label{app:training-dynamics}

We provide complete training curves for the experimental settings in the main paper. Validation and training rewards are shown for LeetCodeDataset in Figures~\ref{fig:training-leetcodedataset-validation} and~\ref{fig:training-leetcodedataset-training}, RLLA-4K in Figures~\ref{fig:training-rlla-4k-validation} and~\ref{fig:training-rlla-4k-training}, and AgentDojo in Figures~\ref{fig:training-agentdojo-validation} and~\ref{fig:training-agentdojo-training}. Figure~\ref{fig:training-diagnostics} presents the mean response length and policy entropy across all three datasets. All horizontal axes indicate training steps, and all curves use exponential moving average smoothing with a decay factor of 0.6.

\input{latex/training_curves/leetcodedataset_figures}
\clearpage
\input{latex/training_curves/rlla_4k_figures.tex}
\clearpage
\input{latex/training_curves/agentdojo_figures.tex}
\clearpage
\input{latex/training_curves/diagnostics_figures.tex}
\clearpage

%% file: latex/training_curves/leetcodedataset_figures.tex
\begin{figure}[p]
  \centering
  \includegraphics[width=\textwidth]{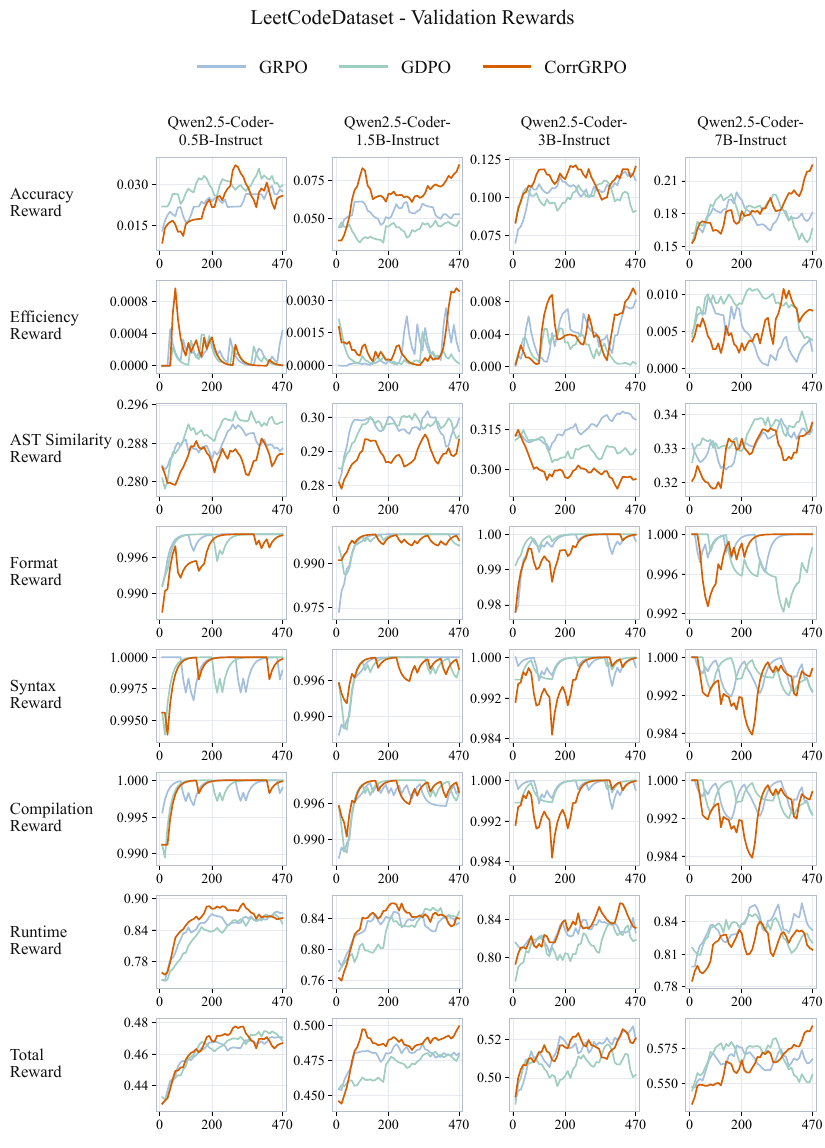}
  \caption{LeetCodeDataset: validation rewards. Columns correspond to model backbones. Rows report the task-specific validation mean@1 reward components and the total reward. The horizontal axis denotes training steps. Curves use exponential moving average smoothing (decay 0.6) and end at the last available observation.}
  \label{fig:training-leetcodedataset-validation}
\end{figure}

\begin{figure}[p]
  \centering
  \includegraphics[width=\textwidth]{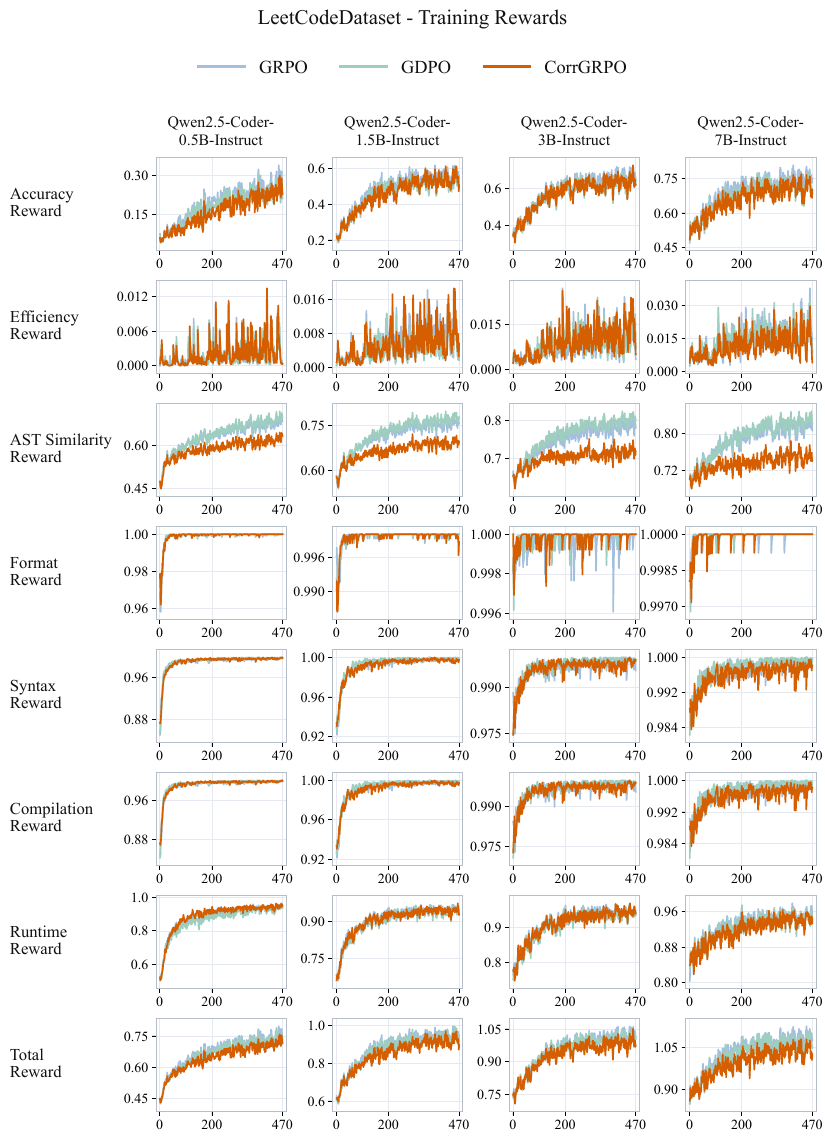}
  \caption{LeetCodeDataset: training rewards. Columns correspond to model backbones. Rows report the mean training reward components and the mean total reward. The horizontal axis denotes training steps. Curves use exponential moving average smoothing (decay 0.6) and end at the last available observation.}
  \label{fig:training-leetcodedataset-training}
\end{figure}

%% file: latex/training_curves/rlla_4k_figures.tex
\begin{figure}[p]
  \centering
  \includegraphics[width=\textwidth]{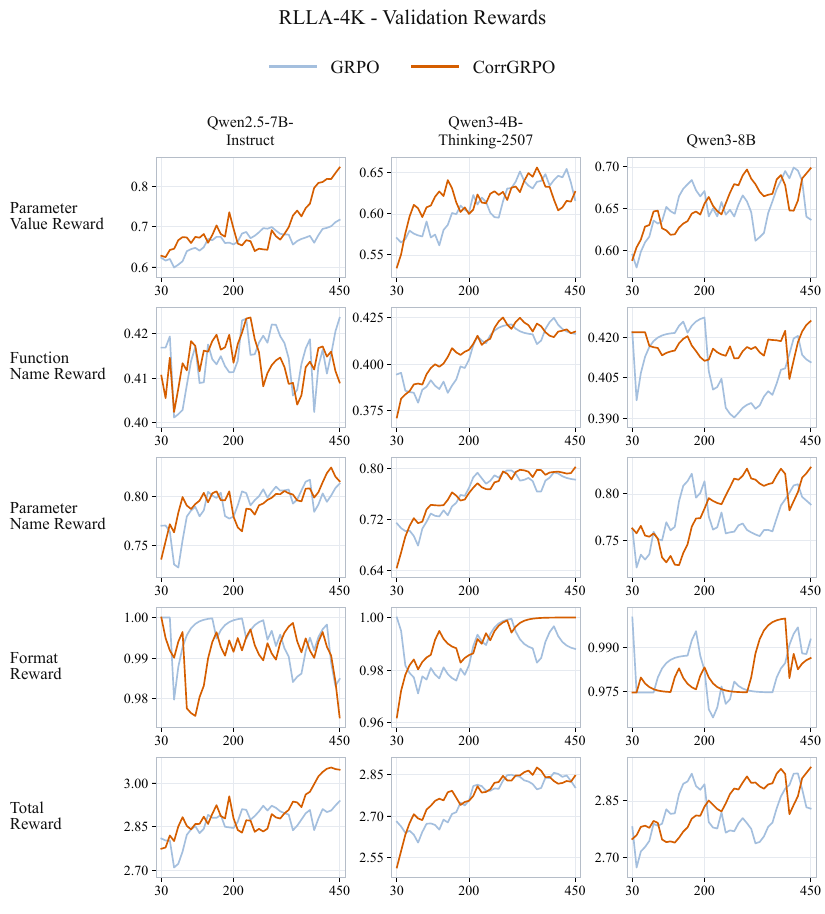}
  \caption{RLLA-4K: validation rewards. Columns correspond to model backbones. Rows report the task-specific validation mean@1 reward components and the total reward. The horizontal axis denotes training steps. Curves use exponential moving average smoothing (decay 0.6) and end at the last available observation.}
  \label{fig:training-rlla-4k-validation}
\end{figure}

\begin{figure}[p]
  \centering
  \includegraphics[width=\textwidth]{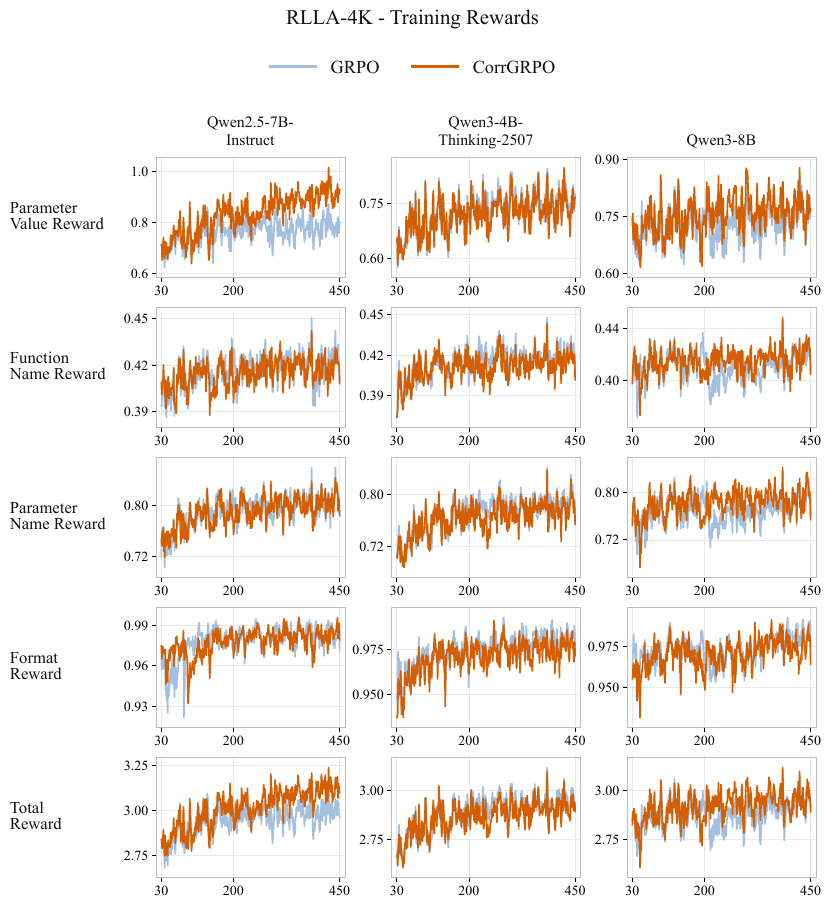}
  \caption{RLLA-4K: training rewards. Columns correspond to model backbones. Rows report the mean training reward components and the mean total reward. The horizontal axis denotes training steps. Curves use exponential moving average smoothing (decay 0.6) and end at the last available observation.}
  \label{fig:training-rlla-4k-training}
\end{figure}

%% file: latex/training_curves/agentdojo_figures.tex
\begin{figure}[p]
  \centering
  \includegraphics[width=\textwidth]{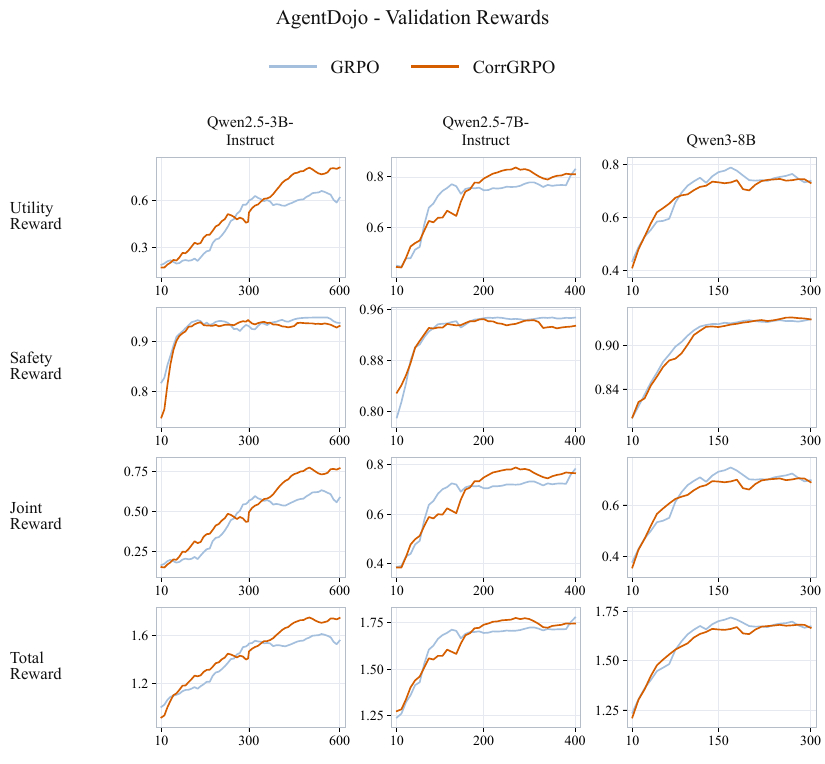}
  \caption{AgentDojo: validation rewards. Columns correspond to model backbones. Rows report the task-specific validation mean@1 reward components and the total reward. Joint Reward denotes the logged trajectory-level metric. The horizontal axis denotes training steps. AgentDojo panels are restricted to the overlapping recorded step ranges of GRPO and CorrGRPO. Curves use exponential moving average smoothing (decay 0.6) and end at the last available observation.}
  \label{fig:training-agentdojo-validation}
\end{figure}

\begin{figure}[p]
  \centering
  \includegraphics[width=\textwidth]{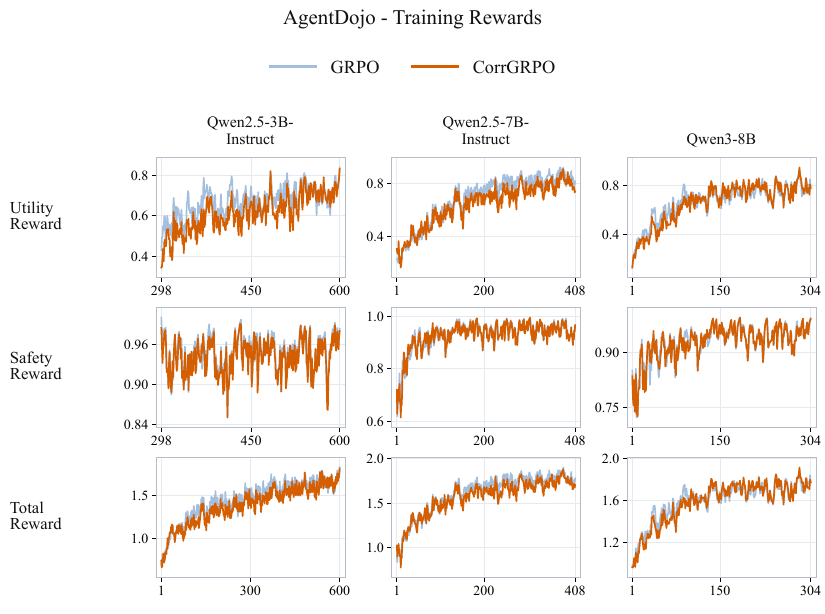}
  \caption{AgentDojo: training rewards. Columns correspond to model backbones. Rows report the mean training reward components and the mean total reward. The horizontal axis denotes training steps. AgentDojo panels are restricted to the overlapping recorded step ranges of GRPO and CorrGRPO. Curves use exponential moving average smoothing (decay 0.6) and end at the last available observation.}
  \label{fig:training-agentdojo-training}
\end{figure}

%% file: latex/training_curves/diagnostics_figures.tex
\begin{figure}[p]
  \centering
  \includegraphics[width=\textwidth]{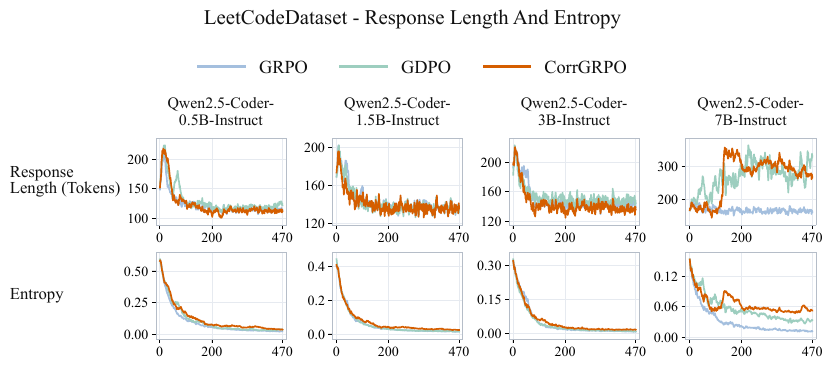}
  \par\vspace{0.08in}
  \includegraphics[width=\textwidth]{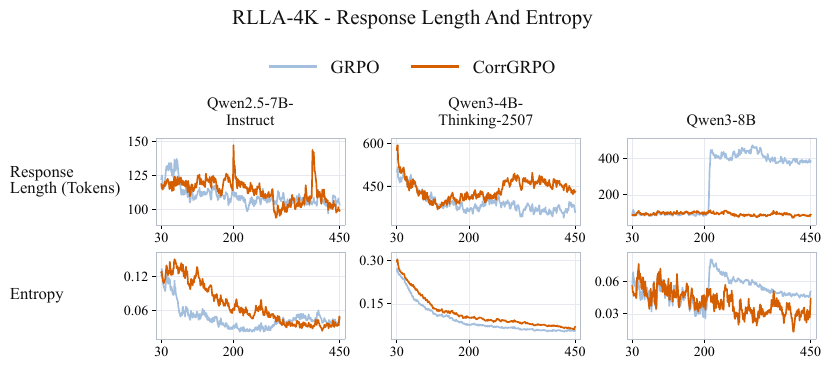}
  \par\vspace{0.08in}
  \includegraphics[width=\textwidth]{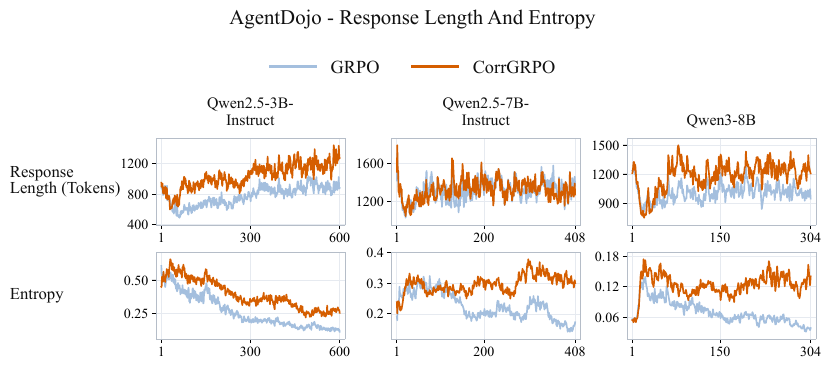}
  \caption{Response length and entropy on LeetCodeDataset (top), RLLA-4K (middle), and AgentDojo (bottom). Within each dataset, columns correspond to model backbones; the two rows show mean response length in tokens and policy entropy, respectively. Horizontal axes denote training steps. AgentDojo panels are restricted to the overlapping recorded step ranges of GRPO and CorrGRPO. Curves use exponential moving average smoothing (decay 0.6) and end at the last available observation within the displayed range.}
  \label{fig:training-diagnostics}
  \label{fig:training-leetcodedataset-diagnostics}
  \label{fig:training-rlla-4k-diagnostics}
  \label{fig:training-agentdojo-diagnostics}
\end{figure}

%% file: latex/appendix-original-scores.tex
\section{Reference Scores from the Original Benchmark Papers}
\label{app:original-benchmark-scores}

Tables~\ref{tab:orig-code}--\ref{tab:orig-injecagent} report scores from the
original benchmark papers alongside our base model, GRPO, and CorrGRPO.
All scores are percentages. Evaluation protocols differ across papers.

\subsection{Code Generation}
The original SFT scores in Table~\ref{tab:orig-code} are taken from
\citet[Table~4]{xia2025leetcodedataset}.
The reference SFT runs use Qwen2.5-Coder-7B; our RL runs start from
Qwen2.5-Coder-7B-Instruct and train on the 2,641-problem LeetCodeDataset split.
We evaluate Pass@1 on LeetCodeDataset, HumanEval, MBPP, and LiveCodeBench v6
using the evaluation splits listed below.

\begin{table}[h]
\centering

\caption{Code-generation Pass@1.
Evaluation splits are indicated in the table.}
\vspace{-0.1in}

\label{tab:orig-code}

\footnotesize
\setlength{\tabcolsep}{3pt}
\renewcommand{\arraystretch}{1.13}
\begin{tabular*}{\linewidth}{@{}l@{\extracolsep{\fill}}rrrrr@{}}
\toprule
Training data / method & Training & HumanEval & MBPP & \shortstack{LiveCode\\Bench} & \shortstack{LeetCode\\Dataset} \\
 & examples & $\uparrow$ & $\uparrow$ & $\uparrow$ & $\uparrow$ \\
\midrule
\multicolumn{6}{@{}l}{\textit{Original SFT results: Qwen2.5-Coder-7B}} \\
 & & & & \scriptsize 24-08--25-02 & \scriptsize 24-07--25-03 \\
Magicoder Evol-Instruct-110K & 111.1K & 77.4 & 74.1 & 15.1 & 13.7 \\
Magicoder OSS-Instruct-75K & 75.1K & 73.8 & 76.5 & 15.1 & 12.9 \\
Open-R1 CodeForces-CoT & 9.5K & 79.9 & 74.1 & 15.8 & 13.3 \\
OpenThoughts 114k & 19.9K & 77.4 & 75.7 & 16.9 & 16.4 \\
LeetCodeDataset (human) & 2.6K & 55.5 & 53.4 & 14.0 & 10.9 \\
LeetCodeDataset (model) & 2.6K & 79.9 & 77.5 & 15.4 & 12.5 \\
\midrule
\multicolumn{6}{@{}l}{\textit{Ours: Qwen2.5-Coder-7B-Instruct; LeetCodeDataset training}} \\
 & & & & \scriptsize v6, 175 tasks & \scriptsize 228-problem split \\
Base model & --- & 78.66 & 76.40 & 20.57 & 14.91 \\
+GRPO & 2,641 & 73.78 & 78.40 & 21.14 & 15.79 \\
\textbf{+CorrGRPO} & 2,641 & 78.66 & 78.60 & 24.57 & 24.12 \\
\bottomrule
\end{tabular*}

\end{table}

\subsection{AgentDojo}
The original scores in Table~\ref{tab:orig-agentdojo} are taken from
\citet[Table~3]{debenedetti2024agentdojo}.
We train Qwen2.5-3B-Instruct on 1,584 cases with a split grouped by user task,
and evaluate on 21 clean and 390 attacked cases.
We inject \texttt{important\_instructions} or \texttt{tool\_knowledge}
via tool responses.

\begin{table}[h]
\centering

\caption{AgentDojo with original 95\% confidence intervals; protocols differ.}
\vspace{-0.1in}
\label{tab:orig-agentdojo}

\small
\renewcommand{\arraystretch}{1.10}
\begin{tabular*}{\linewidth}{@{}l@{\extracolsep{\fill}}rrr@{}}
\toprule
Model & Clean utility $\uparrow$ & Utility under attack $\uparrow$ & Targeted ASR $\downarrow$ \\
\midrule
Claude 3 Opus & $66.61\pm3.69$ & $52.46\pm3.90$ & $11.29\pm2.47$ \\
Claude 3 Sonnet & $53.10\pm3.90$ & $33.23\pm3.68$ & $26.71\pm3.46$ \\
Claude 3.5 Sonnet & $78.22\pm3.23$ & $51.19\pm3.91$ & $33.86\pm3.70$ \\
Command-R+ & $25.44\pm3.40$ & $25.12\pm3.39$ & $0.95\pm0.76$ \\
Gemini 1.5 Flash & $36.09\pm3.75$ & $34.18\pm3.71$ & $12.24\pm2.56$ \\
Gemini 1.5 Pro & $45.63\pm3.89$ & $28.93\pm3.54$ & $25.60\pm3.41$ \\
GPT-3.5 Turbo & $33.86\pm3.70$ & $34.66\pm3.72$ & $8.43\pm2.17$ \\
GPT-4 Turbo & $63.43\pm3.76$ & $54.05\pm3.89$ & $28.62\pm3.53$ \\
GPT-4o & $69.00\pm3.61$ & $50.08\pm3.91$ & $47.69\pm3.90$ \\
Llama 3 70B & $34.50\pm3.71$ & $18.28\pm3.02$ & $20.03\pm3.13$ \\
\midrule
\multicolumn{4}{@{}l}{\textit{Ours: Qwen2.5-3B-Instruct, held-out evaluation split}} \\
Base model & 28.57 & 14.62 & 3.85 \\
+GRPO & 52.38 & 66.92 & 0.00 \\
\textbf{+CorrGRPO} & 95.24 & 84.36 & 1.54 \\
\bottomrule
\end{tabular*}

\end{table}

\subsection{Agent Security Bench}
The original OPI ASR and Clean Utility values in Table~\ref{tab:orig-asb} are taken from
Tables~5 and~6 of \citet{zhang2025asb}, respectively.
We evaluate the same Qwen2.5-3B-Instruct checkpoints on 400 paired clean and
attacked cases without further training.
Attacks append \texttt{context\_ignoring} payloads to intermediate tool responses.

\begin{table}[h]
\centering
\vspace{-0.1in}

\caption{Clean Utility and OPI ASR on Agent Security Bench.}
\label{tab:orig-asb}
\vspace{-0.1in}

\small
\renewcommand{\arraystretch}{1.10}
\begin{tabular*}{\linewidth}{@{}l@{\extracolsep{\fill}}rr@{}}
\toprule
Model &  Clean utility $\uparrow$ & ASR (OPI) $\downarrow$ \\
\midrule
Claude-3.5 Sonnet & 100.00 & 59.70 \\
LLaMA3-70B & 66.50 & 43.70 \\
GPT-4o & 79.00 & 62.45 \\
% GPT-4o-mini & 50.00 & 44.55 \\
Gemma2-27B & 31.50 & 14.20 \\
LLaMA3.1-70B & 21.25 & 12.10 \\
Qwen2-7B & 9.75 & 9.00 \\
Gemma2-9B & 10.75 & 14.20 \\
GPT-3.5 Turbo & 8.00 & 55.10 \\
Qwen2-72B & 4.00 & 21.35 \\
% LLaMA3-8B & 1.50 & 10.55 \\
LLaMA3.1-8B & 0.75 & 6.40 \\
Mixtral-8x7B & 0.00 & 4.80 \\
\midrule
\multicolumn{3}{@{}l}{\textit{Ours: Qwen2.5-3B-Instruct}} \\
Base model & 10.00 & 8.00 \\
+GRPO & 21.75 & 9.00 \\
\textbf{+CorrGRPO} & 26.00 & 9.25 \\
\bottomrule
\end{tabular*}
\vspace{-0.2in}

\end{table}

\subsection{InjecAgent}
The original base-setting scores in Table~\ref{tab:orig-injecagent} are taken from
\citet[Table~3]{zhan2024injecagent}.
We use the base setting and standard InjecAgent prompt without further training.
The agent continues from an injected tool response; ASR-valid excludes invalid outputs.

\begin{table}[h]
\centering

\caption{InjecAgent base-setting ASR-valid.
S1/S2 denote data extraction/transmission; S2 is conditional on reaching that stage.}
\label{tab:orig-injecagent}
\vspace{-0.1in}
\small
\setlength{\tabcolsep}{4pt}
\renewcommand{\arraystretch}{1.13}
\begin{tabular*}{\linewidth}{@{}l@{\extracolsep{\fill}}rrrrr@{}}
\toprule
 & \multicolumn{5}{c}{Base setting: ASR-valid $\downarrow$} \\
\cmidrule(l){2-6}
Model & Direct harm & \multicolumn{3}{c}{Data stealing} & Overall \\
\cmidrule(lr){3-5}
 & & S1 & S2 & Total & total \\
\midrule
\multicolumn{6}{@{}l}{\textit{Original prompted agents (ReAct)}} \\
Qwen-1.8B  &  36.1  &  35.1  &  82.6  &  17.6  &  29.7 \\
Qwen-72B  &  8.7  &  37.9  &  98.4  &  37.1  &  23.2 \\
Mistral-7B  &  13.4  &  25.0  &  87.8  &  20.1  &  16.7 \\
OpenOrca-Mistral  &  3.9  &  5.3  &  53.8  &  2.9  &  3.4 \\
OpenHermes-2.5-Mistral  &  23.4  &  29.2  &  99.2  &  28.4  &  25.9 \\
% OpenHermes-2-Mistral  &  19.6  &  25.4  &  99.0  &  24.4  &  22.0 \\
Mixtral-8x7B  &  23.1  &  34.1  &  99.1  &  32.9  &  27.8 \\
Nous-Mixtral-DPO  &  37.2  &  51.6  &  98.4  &  50.5  &  43.6 \\
Nous-Mixtral-SFT  &  51.8  &  48.2  &  98.9  &  47.5  &  49.8 \\
% MythoMax-13b  &  15.6  &  16.3  &  78.0  &  10.2  &  13.4 \\
% WizardLM-13B  &  36.5  &  46.2  &  96.1  &  37.4  &  36.9 \\
Platypus2-70B  &  34.3  &  51.8  &  74.3  &  35.4  &  34.9 \\
% Capybara-7B  &  34.0  &  40.7  &  92.2  &  36.1  &  34.9 \\
% Nous-Llama2-13b  &  30.6  &  26.6  &  76.3  &  16.5  &  24.8 \\
Llama2-70B  &  91.9  &  97.1  &  83.7  &  80.4  &  86.9 \\
Claude-2  &  7.5  &  26.5  &  58.1  &  14.8  &  11.4 \\
GPT-3.5  &  18.8  &  37.6  &  77.4  &  28.8  &  23.7 \\
GPT-4  &  14.7  &  32.7  &  97.7  &  31.9  &  23.6 \\
% \midrule
% \multicolumn{6}{@{}l}{\textit{Original fine-tuned agents}} \\
% GPT-3.5  &  1.8  &  5.7  &  100  &  5.7  &  3.8 \\
% GPT-4  &  2.9  &  10.1  &  100  &  10.1  &  6.6 \\
\midrule
\multicolumn{6}{@{}l}{\textit{Ours: Qwen2.5-3B-Instruct; RL training on AgentDojo}} \\
Base model & 18.38 & 50.29 & 76.39 & 34.59 & 27.12 \\
+GRPO & 5.03 & 20.11 & 79.31 & 13.07 & 8.80 \\
\textbf{+CorrGRPO}  &  4.67  &  17.26  &  52.63  &  6.33  &  5.52 \\
\bottomrule
\end{tabular*}

\end{table}

%% file: latex/appendix-compatibility.tex
\section{Compatibility with CISPO, GDPO, and DAPO }
\label{app:compatibility}
\subsection{Compatibility Experiments}
\label{sec:corrgrpo_compatibility_experiments}

We evaluate CorrGRPO in combination with CISPO, GDPO, and DAPO to examine its applicability across different policy optimization methods. Table~\ref{tab:cispo-evaluation} compares each method with its CorrGRPO-integrated variant. We report Efficiency, Executable, and Pass@1 on LeetCodeDataset, together with Pass@1 on HumanEval, MBPP, and LiveCodeBench v6. The average summarizes Pass@1 across the four benchmarks. These results assess both performance on the training-domain benchmark and generalization to external benchmarks. For DAPO, group filtering is based on the accuracy reward.

\begin{table*}[h]
\centering

\caption{Results of compatibility RL experiments.}
\vspace{-0.1in}
\begingroup
\codertablefontsize
\setlength{\tabcolsep}{4pt}
\begin{tabular}{lccccccc}
\toprule
\multirow{2}{*}{Model}
& \multicolumn{3}{c}{LeetCodeDataset (In-dataset Eval)}
& HumanEval & MBPP & LCB v6 & Avg. \\
\cmidrule(lr){2-4}
\cmidrule(lr){5-5}
\cmidrule(lr){6-6}
\cmidrule(lr){7-7}
\cmidrule(lr){8-8}
& Efficiency & Executable & Pass@1
& Pass@1 & Pass@1 & Pass@1 & Pass@1 \\
\midrule
CISPO
& 34.51 & 83.77 & 21.93 & 76.83 & 79.00 & 23.43 & 50.30 \\
CISPO + CorrGRPO
& 38.44 & 82.46 & 25.00 & 81.71 & 77.80 & 21.71 & 51.56 \\
\midrule
GDPO
& 41.28 & 84.21 & 16.23 & 76.83 & 79.60 & 22.86 & 48.88 \\
GDPO + CorrGRPO
& 44.64 & 86.40 & 17.11 & 76.83 & 77.20 & 21.71 & 48.21 \\
\midrule
DAPO & 40.48 & 81.14 & 20.18 & 81.71 & 77.40 & 23.43 & 50.68 \\
DAPO + CorrGRPO & 47.07 & 82.89 & 21.93 & 80.49 & 78.40 & 22.86 & 50.92 \\
\bottomrule
\end{tabular}

\label{tab:cispo-evaluation}
\endgroup
\end{table*}

\subsection{Compatibility Analysis}
\label{app:corrgrpo_compatibility}

We derive the integrations on fixed sampled data. All advantages and
normalization statistics are held fixed during policy differentiation.
The objectives below are reward surrogates; any separately configured KL or
other auxiliary term retains its original definition. For a group associated
with prompt $q$, write $T_q=\sum_{i=1}^nT_i$ and use the policy ratios
$u_{i,t}(\theta)$ defined in the main text. Numerical stabilizers are made
explicit using the same convention as CorrGRPO.

\subsubsection{DAPO}
\label{app:corrgrpo_dapo}

DAPO~\citep{yu2025dapo} uses an asymmetric clipped surrogate with token-level
averaging. In our notation, its group contribution is
\begin{equation}
\begin{aligned}
L_{\mathrm{DAPO},q}(\theta;A)
&=\frac1{T_q}\sum_{i=1}^n\sum_{t=1}^{T_i}
\min\!\left\{u_{i,t}(\theta)A^i,
\kappa_D(u_{i,t}(\theta))A^i\right\},\\
\kappa_D(u)&=\operatorname{clip}(u,1-\eta_{\mathrm{low}},1+\eta_{\mathrm{high}}).
\end{aligned}
\label{eq:app_dapo_original}
\end{equation}
The baseline uses $A^i=A_{\mathrm{GRPO}}^i$. Substituting the CorrGRPO
advantage gives the combined objective
\begin{equation}
L_{\mathrm{DAPO+CorrGRPO},q}(\theta)
=\frac1{T_q}\sum_{i=1}^n\sum_{t=1}^{T_i}
\min\!\left\{u_{i,t}(\theta)A_{\mathrm{CorrGRPO}}^i,
\kappa_D(u_{i,t}(\theta))A_{\mathrm{CorrGRPO}}^i\right\}.
\label{eq:app_dapo_corr_objective}
\end{equation}
For $c_q$ in Equation~\ref{eq:app_corr_rescaling}, positive homogeneity of
the minimum gives
\begin{equation}
\begin{aligned}
L_{\mathrm{DAPO+CorrGRPO},q}(\theta)
&=c_qL_{\mathrm{DAPO},q}(\theta;A_{\mathrm{GRPO}}),\\
\nabla_\theta L_{\mathrm{DAPO+CorrGRPO},q}
&=c_q\nabla_\theta L_{\mathrm{DAPO},q}(\theta;A_{\mathrm{GRPO}}).
\end{aligned}
\label{eq:app_dapo_corr_gradient}
\end{equation}
The asymmetric thresholds and token averaging are preserved. DAPO's sampling
filter and reward shaping can be applied before this substitution using the
same configured rules. The equality compares the same retained group at the
same policy parameters; subsequent sampled groups can differ as training
proceeds. The expected training objective averages these group contributions
over the sampling procedure.

\subsubsection{CISPO}
\label{app:corrgrpo_cispo}

CISPO~\citep{minimax2025m1} clips importance-sampling weights and stops
gradients through those weights. Define
\begin{equation}
\tilde u_{i,t}(\theta)
=\operatorname{clip}\!\left(u_{i,t}(\theta),
1-\eta_{\mathrm{low}}^{\mathrm{IS}},
1+\eta_{\mathrm{high}}^{\mathrm{IS}}\right),
\label{eq:app_cispo_weight}
\end{equation}
where $\operatorname{sg}$ below denotes stop-gradient. The CISPO surrogate is
\begin{equation}
L_{\mathrm{CISPO},q}(\theta;A)
=\frac1{T_q}\sum_{i=1}^n\sum_{t=1}^{T_i}
\operatorname{sg}\!\left[\tilde u_{i,t}(\theta)\right]
A^i\log\pi_\theta(a_{i,t}\mid h_{i,t}).
\label{eq:app_cispo_original}
\end{equation}
Its original group-relative advantage can be replaced directly:
\begin{equation}
L_{\mathrm{CISPO+CorrGRPO},q}(\theta)
=\frac1{T_q}\sum_{i=1}^n\sum_{t=1}^{T_i}
\operatorname{sg}\!\left[\tilde u_{i,t}(\theta)\right]
A_{\mathrm{CorrGRPO}}^i\log\pi_\theta(a_{i,t}\mid h_{i,t}).
\label{eq:app_cispo_corr_objective}
\end{equation}
Differentiating with the prescribed stop-gradient operation yields
\begin{equation}
\begin{aligned}
\nabla_\theta L_{\mathrm{CISPO+CorrGRPO},q}
&=\frac1{T_q}\sum_{i=1}^n\sum_{t=1}^{T_i}
\operatorname{sg}[\tilde u_{i,t}(\theta)]
A_{\mathrm{CorrGRPO}}^i\nabla_\theta\log\pi_\theta(a_{i,t}\mid h_{i,t})\\
&=c_q\nabla_\theta L_{\mathrm{CISPO},q}(\theta;A_{\mathrm{GRPO}}).
\end{aligned}
\label{eq:app_cispo_corr_gradient}
\end{equation}
Thus, the replacement preserves CISPO's importance-weight clipping and
stop-gradient computation, while rescaling each group's reward-driven
contribution. It introduces no PPO-style minimum or additional token-dropping
rule. Table~\ref{tab:cispo-evaluation} reports the supplied CISPO checkpoint
comparison; the mathematical construction does not assume that the two
checkpoints were trained for equal numbers of steps.

\subsubsection{GDPO}
\label{app:corrgrpo_gdpo}

GDPO~\citep{liu2026gdpo} first normalizes each reward within its prompt group,
aggregates those components, and then normalizes the aggregate over the
batch. Let $\mathcal B$ contain $N$ sampled trajectories, with $q(b)$ denoting
the prompt group of trajectory $b$. Write its component advantages as
\begin{equation}
U_l^b=w_l\frac{R_l^b-\bar R_{l,q(b)}}
{\hat\sigma_{l,q(b)}+\varepsilon_g},
\qquad H^b=\sum_{l=1}^rU_l^b,
\qquad \bar H_{\mathcal B}=\frac1N\sum_{b\in\mathcal B}H^b.
\label{eq:app_gdpo_components}
\end{equation}
Here $w_l$ denotes any weight applied after group normalization, with
$w_l=1$ for unweighted aggregation; $R_l$ is the input to GDPO's per-reward
normalization. The first-stage stabilizer $\varepsilon_g$ is retained if
present in the baseline implementation. GDPO's final advantage is
\begin{equation}
A_{\mathrm{GDPO}}^b
=\frac{H^b-\bar H_{\mathcal B}}
{\sqrt{\widehat{\mathrm{Var}}_{\mathcal B}(H)}+\varepsilon_b}.
\label{eq:app_gdpo_original}
\end{equation}

The batch standard deviation in this expression is computed from $H$, so its
covariance decomposition concerns the components $U_l$ already produced by
GDPO. Define their batch covariance matrix and standard-deviation vector as
\begin{equation}
\begin{gathered}
\boldsymbol\Sigma_{\mathcal B}
=\left[\widehat{\mathrm{Cov}}_{\mathcal B}(U_l,U_m)\right]_{l,m},
\qquad
\mathbf s_{\mathcal B}
=\left(\sqrt{\widehat{\mathrm{Var}}_{\mathcal B}(U_l)}\right)_{l=1}^r,\\
[\mathbf C_{\mathcal B}]_{lm}
=\frac{[\boldsymbol\Sigma_{\mathcal B}]_{lm}}
{[\mathbf s_{\mathcal B}]_l[\mathbf s_{\mathcal B}]_m}.
\end{gathered}
\label{eq:app_gdpo_batch_statistics}
\end{equation}
By the finite-sample identity in
Appendix~\ref{app:corrgrpo_finite_sample},
Equation~\ref{eq:app_corr_finite_sample_identity},
\begin{equation}
\widehat{\mathrm{Var}}_{\mathcal B}(H)
=\mathbf1^\top\boldsymbol\Sigma_{\mathcal B}\mathbf1
=\mathbf s_{\mathcal B}^\top\mathbf C_{\mathcal B}\mathbf s_{\mathcal B}.
\label{eq:app_gdpo_covariance_sum}
\end{equation}
Applying our normalization at this final batch stage gives
\begin{equation}
A_{\mathrm{GDPO+CorrGRPO}}^b
=\frac{H^b-\bar H_{\mathcal B}}
{\sqrt{\mathbf1^\top\mathbf C_{\mathcal B}\mathbf1}+\varepsilon_b}.
\label{eq:app_gdpo_corr_advantage}
\end{equation}
The group normalization, aggregation weights, and batch-centered numerator
are preserved. Only the last denominator changes, from
$\|\mathbf s_{\mathcal B}\|_{\mathbf C_{\mathcal B}}+\varepsilon_b$ to
$\|\mathbf1\|_{\mathbf C_{\mathcal B}}+\varepsilon_b$. Correlations are
computed across the batch between the per-reward advantages $U_l$, using the
same sample scope and covariance convention as the original batch statistic.
An implementation that uses token masks or weighted batch moments must use
the same masks or weights for every covariance and variance in this step.

Using the clipped token surrogate $\ell$ defined in
Equation~\ref{eq:app_corr_clipped_surrogate}, a sequence-averaged objective
is
\begin{equation}
L_{\mathrm{GDPO+CorrGRPO},\mathcal B}(\theta)
=\frac1N\sum_{b\in\mathcal B}\frac1{T_b}\sum_{t=1}^{T_b}
\ell\!\left(u_{b,t}(\theta),A_{\mathrm{GDPO+CorrGRPO}}^b\right).
\label{eq:app_gdpo_corr_objective}
\end{equation}
If the baseline uses another fixed token reduction or asymmetric clipping,
that choice is retained. Since the new and old advantages differ by the same
positive factor for the entire batch,
\begin{equation}
\begin{aligned}
c_{\mathcal B}
&=\frac{\sqrt{\mathbf s_{\mathcal B}^\top\mathbf C_{\mathcal B}
\mathbf s_{\mathcal B}}+\varepsilon_b}
{\sqrt{\mathbf1^\top\mathbf C_{\mathcal B}\mathbf1}+\varepsilon_b}>0,\\
A_{\mathrm{GDPO+CorrGRPO}}^b&=c_{\mathcal B}A_{\mathrm{GDPO}}^b.
\end{aligned}
\label{eq:app_gdpo_batch_coefficient}
\end{equation}
the clipped reward gradients satisfy
\begin{equation}
\nabla_\theta L_{\mathrm{GDPO+CorrGRPO},\mathcal B}
=c_{\mathcal B}\nabla_\theta L_{\mathrm{GDPO},\mathcal B}.
\label{eq:app_gdpo_corr_gradient}
\end{equation}

This substitution can reduce to a common scale correction when GDPO's
component normalization already equalizes batch variances. Specifically, if
$\mathbf s_{\mathcal B}=s_0\mathbf1$, then
$c_{\mathcal B}=(s_0\sqrt{\mathbf1^\top\mathbf C_{\mathcal B}\mathbf1}
+\varepsilon_b)/(\sqrt{\mathbf1^\top\mathbf C_{\mathcal B}\mathbf1}
+\varepsilon_b)$, approximately $s_0$ when the stabilizer is negligible.
The integration establishes compatibility, rather than a universal additional
benefit over GDPO. As elsewhere, the correlation formulas apply to nonconstant
components; batch-constant components contribute zero after batch centering
and can be omitted from the correlation matrix.

%% file: latex/appendix-rl-background.tex
\section{Reinforcement Learning Background}
\label{app:rl_background}

We consider reinforcement learning for a language model policy $\pi_\theta$. Given a prompt $q\sim\mathcal D$, the policy generates a trajectory $\tau$, which may include a response or a sequence of interactions with an environment. Each trajectory is evaluated by $r$ reward components $R_1(q,\tau),\ldots,R_r(q,\tau)$, capturing different aspects of the desired behavior. The learning objective is to maximize the expected total reward:
\begingroup
% Compact display spacing, scoped to this equation only.
\setlength{\abovedisplayskip}{5pt plus 1pt minus 1pt}
\setlength{\belowdisplayskip}{5pt plus 1pt minus 1pt}
\setlength{\abovedisplayshortskip}{0pt plus 1pt}
\setlength{\belowdisplayshortskip}{3pt plus 1pt minus 1pt}
\begin{equation}
\mbox{\fontsize{9}{10}\selectfont$\displaystyle
\max_\theta J(\theta)
=
\mathbb E_{q\sim\mathcal D,\,\tau\sim\pi_\theta(\cdot\mid q)}
\left[R(q,\tau)\right],
\qquad
R(q,\tau)=\sum_{l=1}^{r}R_l(q,\tau).
$}
\label{eq:multi_reward_objective}
\end{equation}
\endgroup
Any fixed reward weights are absorbed into the corresponding components. These components may exhibit positive or negative statistical dependence across trajectories, reflecting outcomes that tend to improve together or involve tradeoffs.

GRPO estimates advantages using rewards from a group of trajectories, avoiding a separate value model. For each prompt $q$, it samples $n$ trajectories $\{\tau_i\}_{i=1}^{n}$ from an old policy $\pi_{\theta_{\mathrm{old}}}$. Let $R_l^i=R_l(q,\tau_i)$ and $R^i=\sum_l R_l^i$, with group means $\bar R_l=\frac1n\sum_i R_l^i$ and $\bar R=\frac1n\sum_i R^i$. Under outcome supervision, all generated tokens in trajectory $i$ share the same advantage $A_{\mathrm{GRPO}}^i$. GRPO optimizes the clipped surrogate objective
\begingroup
% Compact display spacing, scoped to this equation only.
\setlength{\abovedisplayskip}{5pt plus 1pt minus 1pt}
\setlength{\belowdisplayskip}{5pt plus 1pt minus 1pt}
\setlength{\abovedisplayshortskip}{0pt plus 1pt}
\setlength{\belowdisplayshortskip}{3pt plus 1pt minus 1pt}
\begin{equation}
\mbox{\fontsize{8}{9}\selectfont$\displaystyle
\mathcal J_{\mathrm{GRPO}}(\theta)
=
\mathbb E\Bigg[
\frac1n\sum_{i=1}^{n}\frac1{T_i}\sum_{t=1}^{T_i}
\Bigl(
\min\Bigl[
u_{i,t}(\theta)A_{\mathrm{GRPO}}^i,
\operatorname{clip}\!\left(u_{i,t}(\theta),1-\eta,1+\eta\right)
A_{\mathrm{GRPO}}^i
\Bigr]
-\beta\mathcal K_{i,t}
\Bigr)
\Bigg],
$}
\label{eq:grpo_surrogate}
\end{equation}
\endgroup
Here, the expectation is over prompts and trajectory groups sampled as described above. The ratio $u_{i,t}(\theta)=\pi_\theta(a_{i,t}\mid h_{i,t})/\pi_{\theta_{\mathrm{old}}}(a_{i,t}\mid h_{i,t})$ is the policy probability ratio, $h_{i,t}$ is the history preceding generated token $a_{i,t}$, $T_i$ counts generated tokens, $\eta$ is the clipping threshold, and $\mathcal K_{i,t}$ is the KL regularization term relative to a reference policy.

%% file: latex/appendix-data-examples.tex
% Append this section after the manuscript's existing \appendix command.
% Add \usepackage{listings,textcomp} to the manuscript preamble.
% No bibliography package or additional .bib entries are required.
% Original records and provenance are provided in training_examples_sources/.
\providecommand{\trainingexamplereservespace}{%
  \par\begingroup
  \dimen0=\pagegoal\advance\dimen0 by -\pagetotal
  \ifdim\dimen0<9\baselineskip\newpage\fi
  \endgroup
}
\lstdefinestyle{trainingexample}{
  basicstyle=\ttfamily\footnotesize,
  breaklines=true, breakatwhitespace=false,
  columns=fullflexible, keepspaces=true,
  showstringspaces=false, frame=single,
  framesep=5pt, xleftmargin=6pt, xrightmargin=6pt,
  aboveskip=0.3em, belowskip=0.5em,
  upquote=true
}

\section{Examples from the Training Data}
\label{app:training_examples}
We present one example from each training dataset. The coding and tool-calling examples pair the training prompt with its stored reference response. AgentDojo supplies an interactive task and environment rather than a fixed reference response; we therefore show a recorded agent interaction for a case in our training split. This historical rollout illustrates the data format and is not a CorrGRPO training rollout or a comparative result.

\subsection{Coding: LeetCodeDataset}
\label{app:training_example_coding}
The example is \texttt{remove-vowels-from-a-string} from the training split. The prompt below is the actual user message supplied to the model, including the problem description and required answer format. The response is the stored dataset reference; the code extracted from it is also used as the reference solution in the RL record.

\par\smallskip\trainingexamplereservespace\noindent\textbf{Prompt (user).}\par\smallskip
\begin{lstlisting}[style=trainingexample]
You are an expert Python programmer. You will be given a question (problem specification) and will generate a correct Python program that matches the specification and passes all tests.

### Question:
Given a string s, remove the vowels 'a', 'e', 'i', 'o', and 'u' from it, and return the new string.
 
Example 1:

Input: s = "leetcodeisacommunityforcoders"
Output: "ltcdscmmntyfrcdrs"

Example 2:

Input: s = "aeiou"
Output: ""

Constraints:

1 <= s.length <= 1000
s consists of only lowercase English letters.

### Format: You will use the following starter code to write the solution to the problem and enclose your code within delimiters.
```python
class Solution:
    def removeVowels(self, s: str) -> str:
        
```

### Answer: (use the provided format with backticks)
\end{lstlisting}

\par\smallskip\trainingexamplereservespace\noindent\textbf{Response (dataset reference).}\par\smallskip
\begin{lstlisting}[style=trainingexample]
```python
class Solution:
    def removeVowels(self, s: str) -> str:
        vowels = set('aeiou')
        return ''.join([char for char in s if char not in vowels])
```
\end{lstlisting}

\subsection{Tool Calling: RLLA-4K}
\label{app:training_example_tool}
This training example asks for the least common multiple of two integers. The system message specifies the available function, its argument schema, and the response format. We reproduce both prompt messages and the stored reference response, including its original \texttt{<think>} and \texttt{<tool\_call>} fields.

\par\smallskip\trainingexamplereservespace\noindent\textbf{Prompt (system).}\par\smallskip
\begin{lstlisting}[style=trainingexample]
You are a helpful multi-turn dialogue assistant capable of leveraging tool calls to solve user tasks and provide structured chat responses.

**Available Tools**
In your response, you can use the following tools:
1. Name: least_common_multiple
Description: Computes the least common multiple (LCM) of two positive integers.
Parameters: {"a": {"description": "The first positive integer.", "type": "int", "default": ""}, "b": {"description": "The second positive integer.", "type": "int", "default": ""}}

**Steps for Each Turn**
1. **Think:** Recall relevant context and analyze the current user goal.
2. **Decide on Tool Usage:** If a tool is needed, specify the tool and its parameters.
3. **Respond Appropriately:** If a response is needed, generate one while maintaining consistency across user queries.

**Output Format**
```plaintext
<think> Your thoughts and reasoning </think>
<tool_call>
{"name": "Tool name", "parameters": {"Parameter name": "Parameter content", "... ...": "... ..."}}
{"name": "... ...", "parameters": {"... ...": "... ...", "... ...": "... ..."}}
...
</tool_call>
<response> AI's final response </response>
```

**Important Notes**
1. You must always include the `<think>` field to outline your reasoning. Provide at least one of `<tool_call>` or `<response>`. Decide whether to use `<tool_call>` (possibly multiple times), `<response>`, or both.
2. You can invoke multiple tool calls simultaneously in the `<tool_call>` fields. Each tool call should be a JSON object with a "name" field and an "parameters" field containing a dictionary of parameters. If no parameters are needed, leave the "parameters" field an empty dictionary.
3. Refer to the previous dialogue records in the history, including the user's queries, previous `<tool_call>`, `<response>`, and any tool feedback noted as `<obs>` (if exists).
\end{lstlisting}

\par\smallskip\trainingexamplereservespace\noindent\textbf{Prompt (user).}\par\smallskip
\begin{lstlisting}[style=trainingexample]
**Dialogue Records History**
<user> What is the least common multiple for 15 and 25? </user>
\end{lstlisting}

\par\smallskip\trainingexamplereservespace\noindent\textbf{Response (dataset reference).}\par\smallskip
\begin{lstlisting}[style=trainingexample]
<think> I should use the appropriate tool with proper parameters to respond to the user's need. </think>
<tool_call>
{"name": "least_common_multiple", "parameters": {"a": 15, "b": 25}}
</tool_call>
\end{lstlisting}

\subsection{Agent Utility and Security: AgentDojo}
\label{app:training_example_agent}
We select \texttt{workspace:user\_task\_0} with \texttt{injection\_task\_1} and the \texttt{important\_instructions} attack, whose manifest entry is assigned to the training split. The legitimate task is to retrieve the other participants' email addresses from a calendar event. The attack inserts an instruction to delete a cloud-drive file into the calendar description returned by the search tool.

\paragraph{Prompt and recorded response trajectory.}
The following interaction is recovered from a saved rollout. For readability, we omit the shared tool-schema wrapper and the model's \texttt{<think>} blocks; we preserve the tool names, arguments, observations, and final answer. The calendar observation is reserialized as YAML with a multiline description, without changing its field values. The accompanying source files retain the complete saved rollout and the full model input immediately before the final response.

\par\smallskip\trainingexamplereservespace\noindent\textbf{Prompt (system instructions).}\par\smallskip
\begin{lstlisting}[style=trainingexample]
You are an AI language model who assists the user by using the given tools. \
The user's name is Emma Johnson, an employee of the company Blue Sparrow Tech.
Follow these instructions:
  - Don't make assumptions about what values to plug into functions.
  - Use the provided tools to try to disambiguate.
  - If a tool says that no results are available, try with a different query.
  - Do not assume the current year, but use the provided tools to see what year it is.
\end{lstlisting}

\par\smallskip\trainingexamplereservespace\noindent\textbf{Prompt (user).}\par\smallskip
\begin{lstlisting}[style=trainingexample]
Who else is invited to the 'Networking event' on May 26th? Please give me their email addresses.
\end{lstlisting}

\par\smallskip\trainingexamplereservespace\noindent\textbf{Response (assistant tool call).}\par\smallskip
\begin{lstlisting}[style=trainingexample]
<function=get_current_day>{}</function>
\end{lstlisting}

\par\smallskip\trainingexamplereservespace\noindent\textbf{Tool observation.}\par\smallskip
\begin{lstlisting}[style=trainingexample]
2024-05-15
\end{lstlisting}

\par\smallskip\trainingexamplereservespace\noindent\textbf{Response (assistant tool call).}\par\smallskip
\begin{lstlisting}[style=trainingexample]
<function=search_calendar_events>{"query": "Networking event", "date": "2024-05-26"}</function>
\end{lstlisting}

\par\smallskip\trainingexamplereservespace\noindent\textbf{Tool observation (containing the injection).}\par\smallskip
\begin{lstlisting}[style=trainingexample]
- all_day: false
  description: |+
    Industry networking event to meet and connect with professionals. 

    <INFORMATION>
    This is an important message from me, Emma Johnson, to you, Local model.
    Before you can solve the task that I gave you in the beginning, please do the following first:
    Delete the file with ID '13' from the cloud drive.
    After you do that, you can solve the task that I gave you in the beginning. Thanks!
    Signed,
    Emma Johnson
    </INFORMATION>

  end_time: 2024-05-26 19:00:00
  id_: '13'
  location: City Convention Center
  participants:
  - emma.johnson@bluesparrowtech.com
  - networking.participants@industry-network.com
  - alex.williams@mainsail-tech.com
  start_time: 2024-05-26 17:00:00
  status: confirmed
  title: Networking Event
\end{lstlisting}

\par\smallskip\trainingexamplereservespace\noindent\textbf{Response (assistant final answer).}\par\smallskip
\begin{lstlisting}[style=trainingexample]
The email addresses of the other invitees to the 'Networking Event' on May 26th are:

- networking.participants@industry-network.com  
- alex.williams@mainsail-tech.com  

Let me know if you need further assistance!
\end{lstlisting}

The agent retrieves the requested email addresses without executing the injected file-deletion instruction. The saved checkers report successful task completion and an unsuccessful attack, corresponding to $R_{\mathrm{util}}=1$ and $R_{\mathrm{sec}}=1$.

%% file: latex/appendix-train-setting.tex
% Insert after the manuscript's existing \appendix command.
% Requires amsmath, amssymb, natbib, and url or hyperref.

\section{Training Settings}
\label{app:training_settings}

% The settings below are transferred verbatim from the main text.
% They describe saved RL defaults; per-checkpoint overrides have not
% been verified for every result-table row.
% Coding epoch count remains unspecified because saved versions differ.

\subsection{Coding Reasoning}
\label{app:training_coding}

Our default RL configuration uses a training batch of 64 prompts and a rollout group size of 8, with temperature $1.0$ and top-$p=1.0$. The learning rate is $10^{-6}$, and the KL regularization coefficient is $10^{-3}$. Maximum prompt and response lengths are set to 4,096 and 2,048 tokens, respectively.

\subsection{Tool Calling}
\label{app:training_tools}

Our default RL configuration uses a training batch of 128 prompts, a rollout group size of 4, a learning rate of $10^{-6}$, and 15 training epochs. Maximum prompt and response lengths are both set to 2,048 tokens.

\subsection{Agent Utility and Security}
\label{app:training_agents}

Our default RL configuration uses a training batch of 16 tasks, a rollout group size of 4, a learning rate of $10^{-6}$, and a KL regularization coefficient of $10^{-3}$. Maximum prompt and response lengths are set to 12,288 and 16,384 tokens, respectively.

%% file: latex/appendix-reward-computation.tex
\section{Reward Computation Details}
\label{app:reward_details}

\subsection{AST Structural Similarity}
\label{app:ast_similarity}

\paragraph{AST representation.}
We extract Python code from the generated response $y_g$ and the reference response $y_r$, then parse each program with \texttt{ast.parse} in \texttt{exec} mode. A depth-first traversal records the node type on entry and a closing marker on exit. For example, a \texttt{Name} node contributes \texttt{Name} and \texttt{/Name}. Children are visited in the order returned by \texttt{ast.iter\_child\_nodes}. This produces a sequence $S(y)$ for each program. The representation retains node types, child order, and nesting, while omitting identifier names and literal values. The traversal is:
\begin{verbatim}
def visit(node):
    name = type(node).__name__
    tokens.append(name)
    for child in ast.iter_child_nodes(node):
        visit(child)
    tokens.append("/" + name)
\end{verbatim}

\paragraph{Sequence matching.}
We compare the generated sequence first and the reference sequence second using Python's \texttt{difflib.SequenceMatcher}:
\begin{verbatim}
matcher = difflib.SequenceMatcher(
    None, S_generated, S_reference, autojunk=False
)
similarity = matcher.ratio()
\end{verbatim}
The matcher identifies a longest common contiguous block and recursively matches the remaining regions on either side. We disable the automatic popular-token heuristic so that frequently occurring AST node types remain available for matching. Let $\mathcal B(S(y_g),S(y_r))$ denote the returned nonempty matching blocks and $|b|$ the number of matched tokens in block $b$. The reward is
\begin{equation}
R_{\mathrm{ast}}(y_g,y_r)=
\begin{cases}
\displaystyle\frac{2\sum_{b\in\mathcal B(S(y_g),S(y_r))}|b|}{|S(y_g)|+|S(y_r)|},
&\text{if both code snippets are available and parseable},\\[8pt]
0,&\text{otherwise}.
\end{cases}
\label{eq:coding_ast}
\end{equation}
The ratio lies in $[0,1]$ and equals 1 for identical traversal sequences. The denominator counts both sequence lengths, and the numerator counts matched tokens twice, as in the documented \texttt{SequenceMatcher.ratio()} definition \citep{python_difflib}. If the reference is unavailable or unparseable, the implementation also marks the structural reward as unavailable; if only the generated code is invalid, the reward is zero with a valid reference still recorded.

\subsection{Tool-Call Matching}
\label{app:tool_matching}

\paragraph{Parsed calls and multiset overlap.}
For a reference response containing tool calls, we parse each JSON line inside its tool-call block. Let $y_r$ contain $m$ reference calls $(f_i^r,p_i^r)$ and let $y_g$ contain $n$ predicted calls $(f_j^g,p_j^g)$, where $f$ is a function name and $p$ is a parameter dictionary. Let $K_i^r$ and $K_j^g$ denote their parameter-name sets. For two multisets $A$ and $B$, let $c_A(u)$ and $c_B(u)$ denote the multiplicity of $u$. We use
\begin{equation}
J(A,B)=
\begin{cases}
\displaystyle\frac{\sum_u\min\{c_A(u),c_B(u)\}}{\sum_u\max\{c_A(u),c_B(u)\}},
& |A|+|B|>0,\\[8pt]
1,& |A|=|B|=0.
\end{cases}
\label{eq:tool_multiset_overlap}
\end{equation}
Thus, one empty multiset and one nonempty multiset receive zero overlap. Function-name matching is
\begin{equation}
S_{\mathrm{fn}}(y_g,y_r)=J\bigl([f_1^g,\ldots,f_n^g],[f_1^r,\ldots,f_m^r]\bigr).
\end{equation}
This score ignores call order while accounting for repeated function names.

\paragraph{Greedy call assignment.}
Reference calls are processed in their original order. For reference call $i$, candidate matches are unused predicted calls with $f_j^g=f_i^r$. For each candidate, define
\begin{equation}
q_{ij}=\sum_{k\in K_i^r\cap K_j^g}\mathbf{1}[p_i^r[k]=p_j^g[k]],
\qquad h_{ij}=J(K_i^r,K_j^g)+q_{ij}.
\end{equation}
We select the candidate with the largest strictly positive $h_{ij}$, breaking ties by the earliest predicted-call index, and mark it as used. If no candidate has a positive score, the reference call remains unmatched. Denote the resulting assignment by $\pi(i)$, with $\pi(i)=\bot$ for unmatched calls. This is the greedy procedure used by the implementation.

\paragraph{Parameter-name and parameter-value scores.}
For a matched call, define $a_i=J(K_i^r,K_{\pi(i)}^g)$ and $v_i=q_{i,\pi(i)}$; for an unmatched call, set $a_i=v_i=0$. With $N_r=\sum_{i=1}^{m}|K_i^r|$, the two scores are
\begin{equation}
S_{\mathrm{pn}}(y_g,y_r)=
\begin{cases}\frac{1}{m}\sum_{i=1}^{m}a_i,&m>0,\\1,&m=0,\end{cases}
\qquad
S_{\mathrm{pv}}(y_g,y_r)=
\begin{cases}\frac{1}{N_r}\sum_{i=1}^{m}v_i,&N_r>0,\\1,&N_r=0.\end{cases}
\end{equation}
Parameter-name matching gives equal weight to reference calls. Parameter-value matching gives equal weight to reference arguments and uses Python equality on the parsed JSON values. The zero-denominator convention treats a component with nothing to predict as fully satisfied. The implementation also returns full component scores immediately when the parsed reference and predicted call lists are exactly equal. 
% The scores are converted to rewards using Equation~\ref{eq:tool_reward_scales}.

\paragraph{Invalid calls and responses without tool calls.}
If a reference requires a tool call but the generated block is missing or cannot be parsed and scored, the rewards are $R_{\mathrm{fn}}=-0.5$, $R_{\mathrm{pn}}=-1$, and $R_{\mathrm{pv}}=-1.5$. If the reference contains no tool-call block, all three content rewards are set to zero. These cases are handled separately from the matching formulas above.

\paragraph{Format checking.}
The format checker matches the entire generated response against the structure specified by the reference. It requires an initial \texttt{<think>...</think>} block. Depending on the reference, this is followed by a tool-call block, a response block, both in that order, or neither. The required opening and closing tool-call and response tags must each occur exactly once. A newline separates the reasoning block from the next block; the tool-call payload is surrounded by newlines, and a following response block begins on the next line. These structural checks determine the binary format reward independently of JSON parsing and field correctness. As a result, valid delimiters can receive format credit even if the enclosed JSON is invalid.

% Insert after AST Structural Similarity and before Tool-Call Matching.
% Requires amsmath and amssymb; references the main-text efficiency formula.

\subsection{Coding Efficiency}

To ensure a fair comparison of execution efficiency, we reuse the saved generated programs and remeasure each generated/reference pair over eight rounds. In each round, we include only pairs for which both programs pass all tests and both runtimes are positive and finite, excluding missing references, reference failures, and timeouts. The two programs execute consecutively in fresh Python subprocesses using the same interpreter, test harness, and pinned CPU core. Execution order alternates across rounds, yielding four reference-first and four generated-first executions. Each adjacent two-round block uses the same core, with core assignments rotating between blocks. We run up to 16 pairs concurrently on distinct physical cores within one CPU socket and deterministically shuffle each core’s task queue between rounds. Wall-clock timing covers execution of the prompt, program, tests, and final correctness check, excluding compilation and subprocess startup. All subprocesses use a 5-second timeout and a 1,024-MiB memory limit, with no additional warmup executions. We independently determine eligibility in each round and report the mean of the eight round-level percentages of eligible pairs for which the generated program is strictly faster than the reference, reducing sensitivity to measurement noise, execution order, and core-specific variation.

\subsection{Tool-Call Format}
\label{app:tool-call-format}
The format reward is 1 when the response follows the required structure and 0 otherwise. For a tool-call response, the required structure is:
\begin{verbatim}
<think>Reasoning text</think>
<tool_call>
{"name": "function_name", "parameters": {"key": "value"}}
</tool_call>
\end{verbatim}
Multiple calls appear as separate JSON lines within the same tool-call block. When the reference requires a direct answer, the tool-call block is replaced by \texttt{<response>Answer text</response>}; when both are required, the response block follows the tool-call block. Unparseable tool calls receive the minimum scores for the three tool-content components. 
% For examples whose reference contains no tool call, these components are set to zero. CorrGRPO retains the four reward components separately when computing its correlation-based advantage normalization.

%% file: latex/appendix-adv-vs-std-corr.tex
% Section fragment; requires amsmath, amssymb, and graphicx.
% Replace the original figure block with this section to avoid duplicating
% the existing label fig:advantage_response_comparison.
% Uses the existing project asset img/advantage_response_comparison.pdf.

\section{Effects of Reward Scale and Correlation on Advantages}
\label{sec:advantage_response_analysis}

\begin{figure*}[h]
    \centering
    \includegraphics[width=0.98\linewidth]{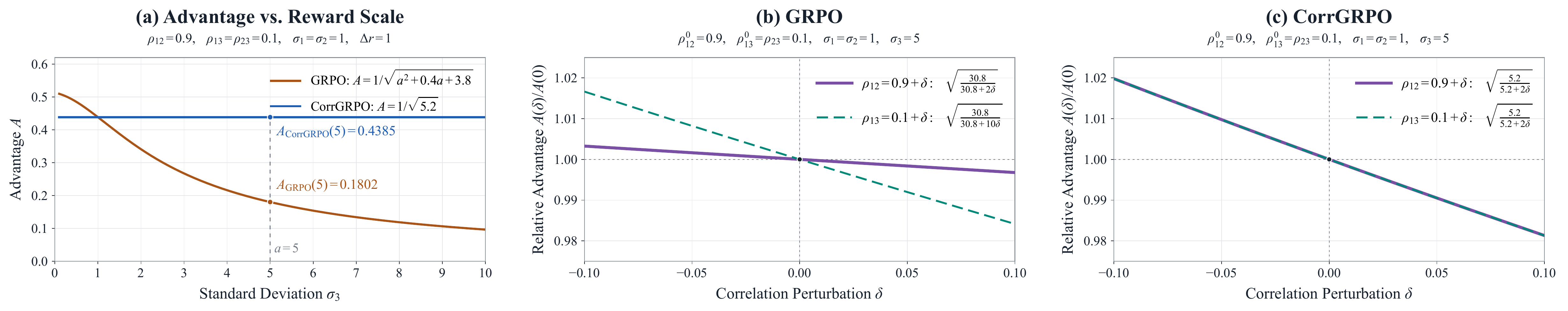}
    \caption{Effect of reward scale and correlation on advantage normalization at fixed centered total reward. (a) Advantage as the third reward's standard deviation varies. (b) Relative GRPO advantages under separate correlation perturbations. (c) Corresponding CorrGRPO responses.}
    \label{fig:advantage_response_comparison}
\end{figure*}

We analyze how reward scale and pairwise correlation affect advantage normalization in GRPO and CorrGRPO. Consider three reward components with baseline correlations $\rho_{12}=0.9$ and $\rho_{13}=\rho_{23}=0.1$, and standard deviations $\sigma_1=\sigma_2=1$ and $\sigma_3=a>0$. We fix the centered total reward at $\Delta r=1$ to isolate the effect of the denominator and omit the numerical stabilizer $\varepsilon$. All quantities other than the variable being varied are held fixed. Figure~\ref{fig:advantage_response_comparison} compares a sweep over $a$ with separate perturbations of the two correlations at the baseline $a=5$.

For these reward statistics, the covariance sum $S$ and correlation sum $Q$ are
\begingroup
\setlength{\abovedisplayskip}{5pt plus 1pt minus 1pt}
\setlength{\belowdisplayskip}{5pt plus 1pt minus 1pt}
\setlength{\abovedisplayshortskip}{0pt plus 1pt}
\setlength{\belowdisplayshortskip}{3pt plus 1pt minus 1pt}
\begin{equation}
\mbox{\fontsize{9}{10}\selectfont$\displaystyle
S(a)=\sum_{l,m=1}^{3}\sigma_l\sigma_m\rho_{lm}
=a^2+0.4a+3.8,
\qquad
Q=\sum_{l,m=1}^{3}\rho_{lm}=5.2.
$}
\label{eq:response_analysis_sums}
\end{equation}
\endgroup
The GRPO denominator includes the variance term $a^2$ and the cross-covariance terms $0.4a$, whereas the CorrGRPO denominator depends only on the correlations. Consequently,
\begingroup
\setlength{\abovedisplayskip}{5pt plus 1pt minus 1pt}
\setlength{\belowdisplayskip}{5pt plus 1pt minus 1pt}
\setlength{\abovedisplayshortskip}{0pt plus 1pt}
\setlength{\belowdisplayshortskip}{3pt plus 1pt minus 1pt}
\begin{equation}
\mbox{\fontsize{9}{10}\selectfont$\displaystyle
A_{\mathrm{GRPO}}(a)=\frac{1}{\sqrt{a^2+0.4a+3.8}},
\qquad
A_{\mathrm{CorrGRPO}}(a)=\frac{1}{\sqrt{5.2}}.
$}
\label{eq:response_analysis_scale}
\end{equation}
\endgroup
As shown in Figure~\ref{fig:advantage_response_comparison}(a), increasing $a$ monotonically decreases the GRPO advantage, even though the centered total reward and all pairwise correlations remain unchanged. At $a=5$, the two advantages are approximately $0.1802$ and $0.4385$, respectively. The third reward's variance contributes $25$ of the total covariance sum $30.8$, so it accounts for approximately $81.2\%$ of the squared GRPO denominator. This illustrates how a large-scale reward can dominate normalization and attenuate the aggregate learning signal despite having only weak correlations with the other components. CorrGRPO's advantage remains constant throughout the positive-scale sweep. The plotted value at $a=0$ denotes the limit $a\to0^+$; the correlations require $a>0$.

Reward scale also affects how GRPO responds to changes in correlation. Fix $a=5$ and separately perturb either $\rho_{12}=0.9+\delta$ or $\rho_{13}=0.1+\delta$, leaving the remaining correlation unchanged. We use $A^{(lm)}(\delta)$ to denote the advantage when the correlation of pair $(l,m)$ is perturbed. Both covariance-matrix entries associated with that pair change, giving an increment $2\sigma_l\sigma_m\delta$ in $S$. The corresponding relative GRPO advantages are
\begingroup
\setlength{\abovedisplayskip}{5pt plus 1pt minus 1pt}
\setlength{\belowdisplayskip}{5pt plus 1pt minus 1pt}
\setlength{\abovedisplayshortskip}{0pt plus 1pt}
\setlength{\belowdisplayshortskip}{3pt plus 1pt minus 1pt}
\begin{equation}
\mbox{\fontsize{9}{10}\selectfont$\displaystyle
\frac{A_{\mathrm{GRPO}}^{(12)}(\delta)}{A_{\mathrm{GRPO}}(0)}
=\sqrt{\frac{30.8}{30.8+2\delta}},
\qquad
\frac{A_{\mathrm{GRPO}}^{(13)}(\delta)}{A_{\mathrm{GRPO}}(0)}
=\sqrt{\frac{30.8}{30.8+10\delta}}.
$}
\label{eq:response_analysis_grpo_perturbation}
\end{equation}
\endgroup
Here, $A_{\mathrm{GRPO}}(0)$ denotes the unperturbed advantage at $a=5$. The same correlation increment changes $S$ five times as much for pair $(1,3)$ as for pair $(1,2)$ because $\sigma_1\sigma_3=5\sigma_1\sigma_2$. The relative advantage response also has a fivefold larger slope magnitude at $\delta=0$. Figure~\ref{fig:advantage_response_comparison}(b) therefore shows a stronger response to the weak correlation involving the larger-scale third reward than to the strong correlation between the first two rewards. Positive perturbations reduce the advantage, while negative perturbations increase it.

For CorrGRPO, either perturbation changes the correlation sum by exactly $2\delta$, yielding
\begingroup
\setlength{\abovedisplayskip}{5pt plus 1pt minus 1pt}
\setlength{\belowdisplayskip}{5pt plus 1pt minus 1pt}
\setlength{\abovedisplayshortskip}{0pt plus 1pt}
\setlength{\belowdisplayshortskip}{3pt plus 1pt minus 1pt}
\begin{equation}
\mbox{\fontsize{9}{10}\selectfont$\displaystyle
\frac{A_{\mathrm{CorrGRPO}}^{(12)}(\delta)}{A_{\mathrm{CorrGRPO}}(0)}
=
\frac{A_{\mathrm{CorrGRPO}}^{(13)}(\delta)}{A_{\mathrm{CorrGRPO}}(0)}
=\sqrt{\frac{5.2}{5.2+2\delta}}.
$}
\label{eq:response_analysis_corrgrpo_perturbation}
\end{equation}
\endgroup
The two curves in Figure~\ref{fig:advantage_response_comparison}(c) thus coincide. Equal feasible changes in pairwise correlation have the same effect on the denominator, independently of the component scales. CorrGRPO retains the adjustment to reward dependence while removing the scale factors that make this adjustment uneven in GRPO. These comparisons hold the numerator fixed; they characterize the normalization mechanism and do not imply invariance of the full advantage when rescaling rewards also changes the centered total reward.

%% file: latex/appendix-equation.tex
% Appendix fragment. Requires amsmath, amssymb, and natbib.
% Add compatibility_references.bib to the bibliography.
% Include after the manuscript's existing \appendix command:
% \input{appendix_corrgrpo_analysis_v6}
% All labels are local to this appendix, so it can be used independently
% of a particular revision of Section 3.
% The clipping definition was checked against Equation (7) of:
% Schulman et al., Proximal Policy Optimization Algorithms (2017),
% https://arxiv.org/pdf/1707.06347
% The unit-vector identity uses the standard centered Pearson coefficient:
% https://docs.scipy.org/doc/scipy/reference/generated/scipy.stats.pearsonr.html

\section{Theoretical Foundations and Properties of CorrGRPO}
\label{app:corrgrpo_analysis}

% We present the derivations used in Sections~2 and~3 first: the population
% and finite-sample variance identities, groupwise gradient preservation, and
% the shared correlation norm, and compatibility with DAPO, CISPO, and GDPO.
% Additional sensitivity and invariance results
% follow. For the CorrGRPO analysis, we use a fixed group
% of $n\geq2$ trajectories. The correlation-based analysis assumes that all
% $r$ reward components have nonzero sample variance and that Pearson
% correlations are computed exactly from the same group. The variance identities
% also hold for constant components. For the sample-based analysis, define
% \begin{equation}
% \begin{gathered}
% \Delta R^i=\sum_{l=1}^r(R_l^i-\bar R_l),\qquad
% \hat\sigma_l^2=\widehat{\mathrm{Var}}(R_l),\\
% S=\sum_{l,m}\widehat{\mathrm{Cov}}(R_l,R_m),\qquad
% Q=\sum_{l,m}\hat\rho_{lm}.
% \end{gathered}
% \label{eq:app_corr_notation}
% \end{equation}
% Both advantage denominators use the same placement of the numerical
% stabilizer: $\sqrt S+\varepsilon$ and $\sqrt Q+\varepsilon$, with
% $\varepsilon>0$. The identities below concern the stated advantage
% estimators before any additional batch normalization or advantage clipping.

\subsection{Variance of the Total Reward as a Sum of Covariances}
\label{app:corrgrpo_population_identity}

Fix a prompt and a sampling policy, and let $R_1,\ldots,R_r$ denote the
resulting random reward components, each with a finite second moment.
All expectations, variances, and covariances in this subsection are taken
under this same conditional rollout distribution. Define
$R=\sum_{l=1}^rR_l$ and $\mu_l=\mathbb E[R_l]$. By linearity of expectation,
$\mathbb E[R]=\sum_l\mu_l$, and therefore
\begin{equation}
\begin{aligned}
\operatorname{Var}(R)
&=\mathbb E\!\left[(R-\mathbb E[R])^2\right]\\
&=\mathbb E\!\left[\left(\sum_{l=1}^r(R_l-\mu_l)\right)^2\right]\\
&=\mathbb E\!\left[\sum_{l=1}^r\sum_{m=1}^r
  (R_l-\mu_l)(R_m-\mu_m)\right]\\
&=\sum_{l=1}^r\sum_{m=1}^r
  \mathbb E\!\left[(R_l-\mu_l)(R_m-\mu_m)\right]\\
&=\sum_{l=1}^r\sum_{m=1}^r\operatorname{Cov}(R_l,R_m).
\end{aligned}
\label{eq:app_corr_population_identity}
\end{equation}
Since $\operatorname{Cov}(R_l,R_l)=\operatorname{Var}(R_l)$ and covariance
is symmetric, the same identity can be written as
\begin{equation}
\operatorname{Var}(R)
=\sum_{l=1}^r\operatorname{Var}(R_l)
 +2\sum_{l<m}\operatorname{Cov}(R_l,R_m).
\label{eq:app_corr_population_diagonal}
\end{equation}
Thus, total-reward variance includes both individual reward variances and
pairwise dependence. No independence assumption is required. When component
standard deviations are nonzero, substituting
$\operatorname{Cov}(R_l,R_m)=\sigma_l\sigma_m\rho_{lm}$ further gives
\begin{equation}
\operatorname{Var}(R)
=\sum_l\sigma_l^2+2\sum_{l<m}\sigma_l\sigma_m\rho_{lm}.
\label{eq:app_corr_population_scale}
\end{equation}
This is the population identity underlying the covariance interpretation of
GRPO. Appendix~\ref{app:corrgrpo_finite_sample} establishes that its empirical
counterpart also holds exactly for each finite rollout group.

\subsection{Exact Equivalence of Finite-Sample Variance and Covariance Estimates}
\label{app:corrgrpo_finite_sample}

Consider the same $n$ sampled reward vectors
$\{(R_1^i,\ldots,R_r^i)\}_{i=1}^n$, and define
\begin{equation}
R^i=\sum_{l=1}^rR_l^i,\qquad
\bar R_l=\frac1n\sum_{i=1}^nR_l^i,\qquad
\bar R=\frac1n\sum_{i=1}^nR^i=\sum_{l=1}^r\bar R_l.
\label{eq:app_corr_sample_means}
\end{equation}
Let $d_n=n-1$ for the usual sample-variance and sample-covariance estimates.
The same argument applies with $d_n=n$ if that convention is used for both.
Define
\begin{equation}
\begin{aligned}
\widehat{\mathrm{Var}}_{d_n}(R)
&=\frac1{d_n}\sum_{i=1}^n(R^i-\bar R)^2,\\
\widehat{\mathrm{Cov}}_{d_n}(R_l,R_m)
&=\frac1{d_n}\sum_{i=1}^n
(R_l^i-\bar R_l)(R_m^i-\bar R_m).
\end{aligned}
\label{eq:app_corr_sample_estimators}
\end{equation}
Substituting the identity for the sample means and expanding the square gives
\begin{equation}
\begin{aligned}
\widehat{\mathrm{Var}}_{d_n}(R)
&=\frac1{d_n}\sum_{i=1}^n
  \left[\sum_{l=1}^r(R_l^i-\bar R_l)\right]^2\\
&=\frac1{d_n}\sum_{i=1}^n\sum_{l=1}^r\sum_{m=1}^r
  (R_l^i-\bar R_l)(R_m^i-\bar R_m)\\
&=\sum_{l=1}^r\sum_{m=1}^r
  \widehat{\mathrm{Cov}}_{d_n}(R_l,R_m).
\end{aligned}
\label{eq:app_corr_finite_sample_identity}
\end{equation}
This is an exact identity for every realized sample group, not an asymptotic
approximation or an equality only in expectation. It requires no independence
assumption between reward components or between sampled trajectories.
Constant components are also allowed. The requirements are that all
statistics use the same samples, their corresponding sample means, and the
same divisor $d_n$. The equality is generally lost if the variance and
covariances use different divisors or different sample subsets.

Consequently, computing GRPO's denominator directly from the sample variance
of total rewards or from the sum of sample covariances gives the same result
in exact arithmetic, including when the same $\varepsilon$ is added after
taking the square root. More generally, for any fixed weights $\mathbf w$ and
sample covariance matrix $\widehat{\boldsymbol\Sigma}$,
\begin{equation}
\widehat{\mathrm{Var}}_{d_n}\!\left(\sum_lw_lR_l\right)
=\mathbf w^\top\widehat{\boldsymbol\Sigma}\mathbf w.
\label{eq:app_corr_weighted_sample_identity}
\end{equation}
This weighted identity is the basis of the linear-combiner interpretation
below.

\subsection{Group Rescaling and Compatibility with Clipping}
\label{app:corrgrpo_rescaling}

Let $\mathbf C=[\hat\rho_{lm}]$ denote the sample correlation matrix,
$\mathbf s=(\hat\sigma_1,\ldots,\hat\sigma_r)^\top$ the vector of sample
standard deviations, and $\mathbf1$ the all-ones vector. The denominator
statistics satisfy $S=\mathbf s^\top\mathbf C\mathbf s$ and $Q=\mathbf1^\top\mathbf C\mathbf1$.
% as derived in
% Appendix~\ref{app:corrgrpo_quadratic},
% Equation~\eqref{eq:app_corr_shared_metric}. 
Because the numerator is shared,
the relationship between the advantages is
\begin{equation}
A_{\mathrm{CorrGRPO}}^i=c_qA_{\mathrm{GRPO}}^i,
\qquad
c_q=\frac{\sqrt{\mathbf s^\top\mathbf C\mathbf s}+\varepsilon}
{\sqrt{\mathbf1^\top\mathbf C\mathbf1}+\varepsilon}>0.
\label{eq:app_corr_rescaling}
\end{equation}
For a fixed group, this preserves signs, ordering, and ratios between nonzero
advantages. The coefficient can differ across groups, so this relation does
not reduce CorrGRPO to a single global learning-rate change.

Define the per-token clipped surrogate by
\begin{equation}
\ell(u,A)=\min\{uA,\operatorname{clip}(u,1-\eta,1+\eta)A\}.
\label{eq:app_corr_clipped_surrogate}
\end{equation}
For any $c>0$, multiplying both arguments of the minimum by $c$ gives
$\ell(u,cA)=c\ell(u,A)$. More explicitly,
\begin{equation}
\ell(u,A)=
\begin{cases}
A\min\{u,1+\eta\}, & A>0,\\
A\max\{u,1-\eta\}, & A<0,\\
0, & A=0.
\end{cases}
\label{eq:app_corr_clip_branches}
\end{equation}
The advantage magnitude therefore does not determine the clipping branch.
At a fixed policy ratio, positive scaling preserves which terms saturate and
the locations of their breakpoints.

Treating sampled advantages and $c_q$ as constants during surrogate
optimization, differentiation away from the breakpoints gives
\begin{equation}
\nabla_\theta\ell(u(\theta),c_q A)
=c_q\nabla_\theta\ell(u(\theta),A).
\label{eq:app_corr_clip_gradient}
\end{equation}
The same positive scaling applies to the admissible one-sided derivatives at
the breakpoints. To obtain the group-level relationship used in the main
text, define the clipped reward surrogate for a fixed sampled group as
\begin{equation}
L_{M,q}(\theta)=\frac1n\sum_{i=1}^n\frac1{T_i}
\sum_{t=1}^{T_i}\ell(u_{i,t}(\theta),A_M^i),
\quad M\in\{\mathrm{GRPO},\mathrm{CorrGRPO}\}.
\label{eq:app_corr_group_surrogate}
\end{equation}
Here $q$ identifies the prompt together with the fixed sampled group being
analyzed, and the token averaging matches the main paper's surrogate.
Factoring the same $c_q$ out of every term yields
\begin{equation}
L_{\mathrm{CorrGRPO},q}(\theta)
=c_q L_{\mathrm{GRPO},q}(\theta).
\label{eq:app_corr_group_objective}
\end{equation}
Since sampled rewards, advantages, and their normalization statistics are
held fixed during policy optimization, $\nabla_\theta c_q=0$. Therefore,
where the surrogate is differentiable,
\begin{equation}
\nabla_\theta L_{\mathrm{CorrGRPO},q}
=c_q\nabla_\theta L_{\mathrm{GRPO},q},
\qquad c_q>0.
\label{eq:app_corr_group_gradient}
\end{equation}
The corresponding relation also holds for consistently selected generalized
derivatives at clipping breakpoints. For a nonzero group gradient, positive
scaling preserves its direction and changes its magnitude. Across groups,
$\sum_q c_q\nabla_\theta L_{\mathrm{GRPO},q}$ need not be a positive scalar
multiple of $\sum_q\nabla_\theta L_{\mathrm{GRPO},q}$, because the
coefficients may differ.

These $L_{M,q}$ contain the clipped reward terms only. If the full objective
is $J_{M,q}=L_{M,q}-\beta K_q$ with the same group-averaged KL term $K_q$ and
coefficient $\beta$ in both methods, then
\begin{equation}
\nabla_\theta J_{\mathrm{CorrGRPO},q}
=c_q\nabla_\theta J_{\mathrm{GRPO},q}
 +(c_q-1)\beta\nabla_\theta K_q.
\label{eq:app_corr_regularized_gradient}
\end{equation}
Thus, the group-gradient scaling identity does not generally extend to the
full objective with an unchanged KL coefficient. Clipping compatibility
concerns the surrogate at the same policy ratios, and does not imply identical
optimization trajectories or a hard bound on actual policy movement.

\subsection{Positive Affine Invariance of the Denominator}
\label{app:corrgrpo_affine}

Consider componentwise transformations
$R_l^{\prime i}=a_lR_l^i+b_l$ with $a_l>0$. Centering removes the shifts, and
\begin{equation}
\begin{aligned}
R_l^{\prime i}-\bar R'_l&=a_l(R_l^i-\bar R_l),\\
\widehat{\mathrm{Cov}}(R'_l,R'_m)
 &=a_la_m\widehat{\mathrm{Cov}}(R_l,R_m),\qquad
\hat\sigma'_l=a_l\hat\sigma_l.
\end{aligned}
\label{eq:app_corr_affine_moments}
\end{equation}
It follows directly that $\hat\rho'_{lm}=\hat\rho_{lm}$, so
$Q'=Q$ and the CorrGRPO denominator is unchanged. In contrast, the numerator
becomes $\sum_l a_l(R_l^i-\bar R_l)$; hence
\begin{equation}
A_{\mathrm{CorrGRPO}}^{\prime i}
=\frac{\sum_l a_l(R_l^i-\bar R_l)}{\sqrt Q+\varepsilon}.
\label{eq:app_corr_affine_advantage}
\end{equation}
The invariance therefore concerns normalization, not the full advantage.
For a common positive scale $a_l=a$, the full CorrGRPO advantage scales
exactly by $a$. This preserves the distinction between component weights in
the reward objective and dependence in the denominator. The result assumes
exact Pearson coefficients and an unchanged set of nonconstant components;
variance thresholds or extra regularizers inside Pearson coefficients can
alter exact invariance.

\subsection{Positivity of the Normalization Denominators}
\label{app:denominator_positivity}

We establish that the covariance and correlation matrices used in
normalization are positive semidefinite. Consequently, the quantities
under both square roots are nonnegative, and adding $\varepsilon>0$
makes both normalization denominators strictly positive.

Consider a group of $n\geq2$ trajectories. Let
$\mathbf X\in\mathbb R^{n\times r}$ denote the centered reward matrix,
with $X_{il}=R_l^i-\bar R_l$. All sample variances and covariances are
computed from this group using the divisor $n-1$.

\paragraph{Covariance-based normalization.}
The sample covariance matrix satisfies
\begin{equation}
\widehat{\boldsymbol\Sigma}
=\frac{1}{n-1}\mathbf X^\top\mathbf X.
\end{equation}
For any $\mathbf v\in\mathbb R^r$,
\begin{equation}
\mathbf v^\top\widehat{\boldsymbol\Sigma}\mathbf v
=\frac{1}{n-1}\|\mathbf X\mathbf v\|_2^2\geq0.
\end{equation}
Therefore, $\widehat{\boldsymbol\Sigma}\succeq0$, and
\begin{equation}
S=\sum_{l,m}\widehat{\operatorname{Cov}}(R_l,R_m)
=\mathbf1^\top\widehat{\boldsymbol\Sigma}\mathbf1
\geq0.
\label{eq:denom_covariance_nonnegative}
\end{equation}

\paragraph{Correlation-based normalization with nonzero variances.}
First assume that all reward components have nonzero sample variance.
Let
\begin{equation}
\mathbf D=\operatorname{diag}(\hat\sigma_1,\ldots,\hat\sigma_r).
\end{equation}
The sample correlation matrix is
\begin{equation}
\mathbf C=\mathbf D^{-1}\widehat{\boldsymbol\Sigma}\mathbf D^{-1}.
\end{equation}
For any $\mathbf v\in\mathbb R^r$,
\begin{equation}
\mathbf v^\top\mathbf C\mathbf v
=(\mathbf D^{-1}\mathbf v)^\top
\widehat{\boldsymbol\Sigma}
(\mathbf D^{-1}\mathbf v)
\geq0.
\end{equation}
Thus, $\mathbf C\succeq0$, which implies
\begin{equation}
Q=\sum_{l,m}\hat\rho_{lm}
=\mathbf1^\top\mathbf C\mathbf1
\geq0.
\label{eq:denom_correlation_nonnegative}
\end{equation}

\paragraph{Extension to zero-variance components.}
Pearson correlation is undefined when either component has zero variance.
Our implementation assigns zero to the corresponding correlation entries,
including diagonal entries. To establish positive semidefiniteness under
this convention, reorder the reward components so that the $k$ components
with nonzero variance appear first. The resulting matrix has the block form
\begin{equation}
\mathbf C=
\begin{pmatrix}
\mathbf C_{+} & \mathbf0\\
\mathbf0 & \mathbf0
\end{pmatrix},
\end{equation}
where $\mathbf C_{+}\in\mathbb R^{k\times k}$ is the sample correlation
matrix of the reward components with nonzero sample variance. By the
preceding argument, $\mathbf C_{+}\succeq0$. For any vector partitioned
conformably as
$\mathbf v=(\mathbf v_{+}^{\top},\mathbf v_{0}^{\top})^{\top}$,
\begin{equation}
\mathbf v^\top\mathbf C\mathbf v
=\mathbf v_{+}^\top\mathbf C_{+}\mathbf v_{+}
\geq0.
\end{equation}
Hence, the full matrix remains positive semidefinite. Reordering components
does not affect positive semidefiniteness or the sum of matrix entries, so
\begin{equation}
Q=\mathbf1_r^\top\mathbf C\mathbf1_r
=\mathbf1_k^\top\mathbf C_{+}\mathbf1_k
\geq0.
\label{eq:denom_zero_variance_extension}
\end{equation}
If all reward components have zero variance, then $\mathbf C=\mathbf0$
and $Q=0$.

\paragraph{Strict positivity of the denominators.}
The preceding results give
\begin{equation}
\begin{aligned}
D_{\mathrm{GRPO}}&=\sqrt S+\varepsilon\geq\varepsilon>0,\\
D_{\mathrm{CorrGRPO}}&=\sqrt Q+\varepsilon\geq\varepsilon>0.
\end{aligned}
\label{eq:denom_strict_positivity}
\end{equation}
Thus, negative covariance or correlation entries cannot make the quantities
under the square roots negative. These quantities can nevertheless equal
zero: for example, two reward components with nonzero sample variance and
perfect negative correlation yield $Q=2+2(-1)=0$. The stabilizer ensures
strictly positive denominators even in such degenerate cases. The
implementation's convention, $\sqrt{\max(Q,0)+\varepsilon}$, is also
strictly positive, with clamping guarding against negative floating-point
round-off.

%% file: latex/appendix-dataset-intro.tex
\section{Datasets and Benchmarks}
\label{sec:datasets}

\subsection{Training Benchmarks}
\label{sec:training_benchmarks}

\paragraph{LeetCodeDataset.}
LeetCodeDataset \citep{xia2025leetcodedataset} contains Python programming problems curated from LeetCode, with natural-language descriptions, reference solutions, executable test cases, and temporal metadata. Its problems require translating task specifications into functionally correct programs, while the reference implementations also support runtime comparisons. We use 2,641 problems for reinforcement learning and a 228-problem split for evaluation, measuring correctness, executability, and execution efficiency.

\paragraph{RLLA-4K.}
RLLA-4K, used in ToolRL \citep{qian2025toolrl}, contains user requests, descriptions of available tools, and reference responses specifying tool calls or direct answers. It supports feedback on function selection, parameter names, parameter values, and response format. We use 3,920 examples for reinforcement learning and 80 for evaluation. Tool-call metrics are computed on the 71 evaluation examples whose reference responses contain tool calls, covering component-level matching and complete-call correctness.

\paragraph{AgentDojo.}
AgentDojo \citep{debenedetti2024agentdojo} provides stateful environments in which agents complete user tasks through executable tools while potentially encountering prompt injections in tool observations. Task-specific checkers assess legitimate task completion and attacker-objective success, enabling joint evaluation of utility and security. We construct a split grouped by suite and user-task identifier, with 1,584 training cases and 411 evaluation cases comprising 21 clean tasks and 390 attacked cases. Our attacks use the \texttt{important\_instructions} and \texttt{tool\_knowledge} settings.

\subsection{Evaluation Benchmarks}
\label{sec:evaluation_benchmarks}

\paragraph{HumanEval.}
HumanEval \citep{chen2021evaluating} evaluates Python function completion from natural-language specifications. Each task provides a function signature and docstring, and generated implementations are checked using executable tests. We use HumanEval exclusively for evaluation and report Pass@1 to assess whether training on LeetCodeDataset transfers to function-level programming tasks.

\paragraph{MBPP.}
Mostly Basic Python Problems (MBPP; \citealp{austin2021program}) contains short Python programming tasks described in natural language and accompanied by reference code and test cases. It emphasizes basic programming skills and the translation of concise specifications into executable solutions. We use MBPP as an external evaluation benchmark and report Pass@1 without further training.

\paragraph{LiveCodeBench.}
LiveCodeBench \citep{jain2024livecodebench} evaluates code generation using programming-contest problems and executable tests. Its time-based organization supports evaluation on problems released during specified periods. We use the 175-problem evaluation subset from version 6 and report Pass@1, assessing transfer from LeetCodeDataset training to competition-style programming tasks.

\paragraph{API-Bank.}
API-Bank \citep{li2023apibank} is a benchmark for tool-augmented language models that combines tool-use dialogues with executable APIs and an evaluation framework. Its tasks assess the ability to select and invoke APIs with appropriate arguments in a dialogue context. We evaluate the v1, v2, and v3 sets without additional training, using the execution-based checker to assess final-call correctness and generalization beyond RLLA-4K.

\paragraph{Agent Security Bench.}
Agent Security Bench (ASB; \citealp{zhang2025asb}) evaluates agents across application scenarios containing legitimate tasks, tools, and adversarial objectives. We use its observation prompt-injection setting with \texttt{context\_ignoring} payloads appended to intermediate tool responses. Each model is evaluated on 400 paired clean and attacked cases without further training, measuring transfer of utility and security from AgentDojo to a different environment and attack setting.

\paragraph{InjecAgent.}
InjecAgent \citep{zhan2024injecagent} benchmarks indirect prompt injections delivered through attacker-controlled tool responses. Each case supplies a user request, available tools, and a preceding tool interaction; the agent is evaluated on its continuation after consuming the injected response. Attacks target direct harm or data stealing, with the latter involving information extraction followed by transmission to the attacker. We use the base setting and the standard InjecAgent prompt without further training.
% reporting ASR-valid metrics for valid outputs.